\pdfoutput=1
\documentclass[11pt]{article}

\usepackage[table,xcdraw]{xcolor}
\usepackage{tcolorbox}

\definecolor{MyBlue}{RGB}{0, 76, 153}

\tcbset{
  colback=gray!5,
  colframe=black!25,
  coltitle=MyBlue,
  colbacktitle=white,
  fonttitle=\bfseries,
  boxrule=0.5pt,
  arc=2pt,
  left=4pt, right=4pt, top=4pt, bottom=4pt,
  boxsep=2pt
}

\usepackage{graphicx}
\usepackage{multirow}
\usepackage{xcolor}
\usepackage{amsmath}
\usepackage{caption}
\usepackage[normalem]{ulem}
\usepackage{textcomp}

\usepackage{booktabs}

\useunder{\uline}{\ul}{}

\usepackage[final]{acl}

\usepackage{times}
\usepackage{latexsym}

\usepackage[T1]{fontenc}

\usepackage[utf8]{inputenc}

\usepackage{microtype}

\usepackage{inconsolata}

\title{\textsc{Strategy-Induct}: Task-Level Strategy Induction\\for Instruction Generation}

\author{
    Po-Chun Chen\textsuperscript{1} \quad    
    Hen-Hsen Huang\textsuperscript{2}\quad
    Hsin-Hsi Chen\textsuperscript{1,3}
    \\
    \textsuperscript{1}Department of Computer Science and Information Engineering, 
    \\ National Taiwan University, Taiwan
    \\
    \textsuperscript{2}Institute of Information Science, Academia Sinica, Taiwan
    \\
    \textsuperscript{3}AI Research Center (AINTU), National Taiwan University, Taiwan
    \\
    \texttt{pcchen@nlg.csie.ntu.edu.tw,}
    \\
    \texttt{hhhuang@iis.sinica.edu.tw, \quad hhchen@ntu.edu.tw}
}

\begin{document}
\maketitle

\begin{abstract}
Designing effective task-level prompts is crucial for improving the performance of Large Language Models (LLMs). While prior work on instruction induction demonstrates that LLMs can infer better instructions with limited examples, existing approaches often rely on input-output pairs, where obtaining labeled answers can be difficult or costly.
To address this limitation, we propose \textsc{Strategy-Induct}, a framework that derives task-level instructions solely from a small set of example questions without requiring labeled answers. Our approach first prompts the model to generate explicit reasoning strategies for each question, forming (strategy, question) pairs. These pairs are then used to induce a task instruction that guides reasoning.
Experiments across multiple tasks and model scales demonstrate that \textsc{Strategy-Induct} outperforms state-of-the-art methods in question-only settings. 
Furthermore, we observe that jointly utilizing LLMs and Large Reasoning Models across task instruction generation and inference may lead to further performance improvements.
\end{abstract}

\begin{figure*}[h!]
    \centering
    \includegraphics[width=\textwidth]{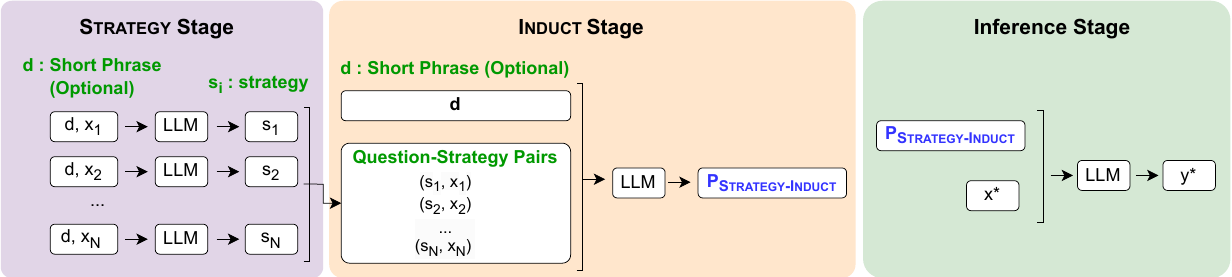}
    \caption{Our proposed \textsc{Strategy-Induct} framework for strategy-based instruction induction. 
    The framework consists of three stages: (1) \textbf{\textcolor[rgb]{0.5,0.3,0.7}{Strategy Stage}}, where the LLM generates strategies ($s_i$) for given inputs ($x_i$). 
    (2) \textbf{\textcolor[rgb]{0.9,0.5,0.0}{INDUCT Stage}}, where question-strategy pairs ($ (s_i, x_i) $) are combined with meta prompts and short phrases to generate an induced prompt ($ P_{\text{\textsc{Strategy-Induct}}} $). 
    (3) \textbf{\textcolor[rgb]{0.1,0.6,0.1}{Inference Stage}}, where $ P_{\text{\textsc{Strategy-Induct}}} $ is used in inference to guide the LLM in solving a new question (x*) in the same task.}
    \label{fig:flow_chart}
\end{figure*}

\section{Introduction}

Large language models (LLMs) have achieved remarkable progress in natural language processing, demonstrating strong capabilities in understanding, generating, and reasoning \cite{Brown2020LanguageMA, Wei2022EmergentAO, chen-etal-2023-self, Wei2023LargerLM, Achiam2023GPT4TR}. 
Providing high-quality task instructions is essential for enabling LLMs to effectively perform tasks, as well-designed instructions can significantly enhance accuracy and consistency. 
However, designing high-quality task instructions is challenging, as it often requires domain expertise and extensive manual effort. 

\citet{honovich-etal-2023-instruction} showed that LLMs can induce task instructions from input-output pairs, a process known as \textit{instruction induction}.  
Building on this concept, subsequent studies have explored automated task instruction generation \cite{Zhou2022LargeLM, Chen2023InstructZeroEI, hsieh-etal-2024-automatic}.  
While effective, these methods often rely on large external resources or an initial instruction prompt, making them less practical for users with limited data or prompt-writing skills \cite{Desmond2024ExploringPE}.  
Moreover, their optimized instructions are often model-specific, limiting generalization across architectures.

To address these limitations, recent studies have explored low-cost instruction induction methods that require only a small number of examples \cite{chen-etal-2024-induct, Zhou2024SelfDiscoverLL, aswani-etal-2024-auto}, allowing LLMs to induce effective task instructions to improve inference performance. While \citet{chen-etal-2024-induct} demonstrates that LLMs can generate better instructions using limited examples and even exhibit cross-model adaptability, their approach still heavily relies on input-output pairs. 
In real-world applications, such reliance may be impractical, as obtaining labeled answers can often be difficult or costly.

To overcome these challenges, we propose \textsc{Strategy-Induct}, a framework that generates task instructions solely from input questions, eliminating the need for labeled answers. Instead of relying on input-output pairs, it creates strategies for each example question and induces a task-level instruction from these strategy-question pairs.

As mentioned earlier, general users often lack the prompt-writing skills needed to communicate effectively with LLMs, making it challenging to craft well-formed and complete instructions.
To address this, we adopt the Short Phrase Prompting concept proposed by \citet{chen-etal-2024-induct}, where a brief task description helps convey task intent. A Short Phrase can be used alongside a minimal user query when needed, or omitted if the question is already self-explanatory. The specific Short Phrases used in our experiments are listed in Appendix~\ref{appendix:dataset_detail}.

\textsc{Strategy-Induct} provides several advantages. First, it enables instruction induction in question-only settings, addressing the critical challenge of missing labeled outputs. Second, unlike prompt optimization approaches, \textsc{Strategy-Induct} generates instructions that generalize across different models without requiring task-specific fine-tuning.

We evaluate \textsc{Strategy-Induct} in question-only scenarios, where each task provides a set of example questions without answers. Our experiments cover multiple datasets, including BBH-Induct, Evals-Induct, and the Shift Cipher Task. Our results show that \textsc{Strategy-Induct} surpasses existing SOTA methods. 
Furthermore, we analyze its effectiveness across different LLM families (e.g., Llama, Mistral, GPT, Gemini), model scales, and model types, including both LLMs and Large Reasoning Models (LRMs), such as GPT o3 mini, demonstrating its scalability and robustness.

\section{\textsc{Strategy-Induct} Framework}

Existing instruction induction methods such as 
INDUCT-LEARN~\cite{chen-etal-2024-induct} require labeled 
input-output pairs to induce task instructions. However, obtaining 
correct answers can be costly or infeasible in many practical 
scenarios. \textsc{Strategy-Induct} removes this requirement by 
operating in a question-only setting. The key enabler is the 
\textsc{Strategy} stage, which generates reasoning strategies for 
each question, providing the structured signal that labeled outputs 
would otherwise supply for the subsequent induction step.

In this section, we formally describe \textsc{Strategy-Induct}, a 
framework for task-level strategy induction that induces task 
instructions from input questions without requiring labeled answers. 
It consists of three stages: Strategy, Induct, and Inference. The 
following subsections define each stage (see 
Figure~\ref{fig:flow_chart} for an overview).

\subsection{\textsc{Strategy} Stage}

The goal of this stage is to generate a reasoning strategy $s_i$ for each input question $x_i$. Given a set of $N$ input questions from the \textbf{same task}:

\begin{equation}
    \mathcal{X} = \{ x_1, x_2, \dots, x_N \}
\end{equation}

Each $x_i$ is a task-specific question. 
The LLM is prompted with a meta prompt $P_{\text{S}}$, the question $x_i$, and optionally a task-level Short Phrase $d$ (a brief task description) to produce the corresponding reasoning strategy:

\begin{equation}
    s_i = \text{LLM}(P_{\text{S}}, d, x_i)
\end{equation}

Repeating this process for each $x_i \in \mathcal{X}$ produces a set of strategy-question pairs:

\begin{equation}
    \mathcal{S} = \{ (s_1, x_1), (s_2, x_2), \dots, (s_N, x_N) \}
\end{equation}

The next stage takes $\mathcal{S}$ as input to induce a task-level instruction prompt.

\subsection{\textsc{Induct} Stage}

In this stage, we construct an induced prompt $P_{\text{\textsc{Strategy-Induct}}}$ using the collected strategy-question pairs. 
Specifically, the LLM is prompted with a meta prompt $P_{\text{I}}$, the Short Phrase $d$, and the set $\mathcal{S}$ to induce a task-level instruction.

\begin{equation}
    P_{\textsc{Strategy-Induct}} = \text{LLM}(P_{\text{I}}, d, \mathcal{S})
\end{equation}

The resulting $P_{\text{\textsc{Strategy-Induct}}}$ is a reusable task-level prompt applicable to other questions within the same task.

\subsection{Inference Stage}

At inference time, the induced prompt $P_{\textsc{Strategy-Induct}}$ is used to solve new questions within the same task. Given a new question $x^*$, the LLM generates a response:

\begin{equation}
    y^* = \text{LLM}(P_{\textsc{Strategy-Induct}}, x^*)
\end{equation}

where $y^*$ is the model's response. Thus, $P_{\textsc{Strategy-Induct}}$ can be reused across questions within the same task.

\section{Experiments}

\subsection{Models}
We conduct experiments using 18 models, spanning four LLM series and two LRMs, covering both open-source and proprietary models. Further details on model versions and configurations are provided in Appendix~\ref{sec:exp_config}.

\begin{table*}[]
\small
\centering
\setlength{\tabcolsep}{3.8pt}
\begin{tabular}{c|cccc|cccc|cccc}
\hline
 & \multicolumn{4}{c|}{BBH-Induct} & \multicolumn{4}{c|}{Evals-Induct} & \multicolumn{4}{c}{Shift Cipher} \\ \hline
Model & ZCoT & SCoT & Induct & Ours & ZCoT & SCoT & Induct & Ours & ZCoT & SCoT & Induct & Ours \\ \hline
\multicolumn{1}{c|}{Llama 3.1 8B} & 62.03 & 56.29 & 59.48 & \textbf{65.33} & 33.80 & 32.09 & 36.09 & \textbf{37.36} & 0.80 & 0.64 & \textbf{2.56} & \textbf{2.56} \\
\multicolumn{1}{c|}{Llama 3.1 70B} & 82.09 & 84.52 & 86.03 & \textbf{88.99} & 55.91 & 56.99 & \textbf{59.45} & 58.91 & 30.40 & 32.16 & 47.84 & \textbf{50.24} \\
\multicolumn{1}{c|}{Llama 3.1 405B} & 85.04 & 83.94 & \textbf{89.28} & 88.52 & 61.57 & 61.48 & \textbf{62.44} & 61.84 & 44.16 & 34.24 & \textbf{65.92} & 61.44 \\
\multicolumn{1}{c|}{Llama 3.3 70B} & 83.59 & 84.52 & \textbf{88.52} & 87.30 & 59.99 & 59.34 & 58.49 & \textbf{60.32} & 34.24 & 31.84 & 53.44 & \textbf{56.00} \\
\hline
\multicolumn{1}{c|}{Mistral Nemo 12B} & \textbf{56.46} & 53.97 & 52.29 & 55.71 & 31.81 & 26.33 & \textbf{35.93} & 35.29 & 0.00 & 0.16 & 0.00 & \textbf{0.48} \\
\multicolumn{1}{c|}{Mistral Small 2} & 70.14 & 69.91 & 69.57 & \textbf{70.32} & 48.34 & 48.18 & \textbf{51.03} & 48.95 & 1.76 & \textbf{2.08} & 0.96 & 1.76 \\
\multicolumn{1}{c|}{Mistral Small 3} & 75.07 & 76.52 & 76.64 & \textbf{80.41} & 53.94 & 56.50 & 55.13 & \textbf{57.22} & 18.24 & 18.08 & 20.48 & \textbf{21.44} \\
\multicolumn{1}{c|}{Mistral Large 2} & 82.03 & 81.86 & 85.57 & \textbf{85.97} & \textbf{63.74} & 58.93 & 63.44 & 60.08 & 38.56 & \textbf{44.32} & 23.84 & 41.12 \\
\hline
\multicolumn{1}{c|}{Gemini 1.5 Flash 8B} & 67.42 & \textbf{69.80} & 67.07 & 69.33 & 44.97 & \textbf{45.24} & 39.25 & 43.20 & 3.52 & \textbf{4.16} & 3.84 & 3.84 \\
\multicolumn{1}{c|}{Gemini 1.5 Flash} & 78.72 & 77.86 & 77.68 & \textbf{81.80} & 53.57 & 55.78 & 54.54 & \textbf{56.56} & 6.88 & 13.76 & 15.68 & \textbf{26.40} \\
\multicolumn{1}{c|}{Gemini 1.5 Pro} & 80.87 & 82.03 & 83.71 & \textbf{85.39} & 64.61 & 64.83 & 63.07 & \textbf{67.18} & 30.40 & 41.28 & \textbf{64.16} & 57.60 \\
\multicolumn{1}{c|}{Gemini 2.0 Flash Lite} & 76.64 & 78.23 & 76.52 & \textbf{81.80} & 60.48 & 59.52 & 60.54 & \textbf{62.49} & 36.16 & 19.68 & \textbf{52.00} & \textbf{52.00} \\
\multicolumn{1}{c|}{Gemini 2.0 Flash} & 78.33 & 79.62 & 79.30 & \textbf{85.45} & 61.35 & 64.22 & 60.01 & \textbf{66.38} & 54.24 & 53.44 & 65.60 & \textbf{67.04} \\
\multicolumn{1}{c|}{Gemini 2.0 Pro Exp} & 83.62 & 85.04 & 80.23 & \textbf{86.72} & 69.69 & 68.88 & 64.47 & \textbf{71.08} & 55.04 & 54.56 & \textbf{77.12} & 71.04 \\
\hline
\multicolumn{1}{c|}{GPT-4o mini} & 79.54 & 80.06 & 78.78 & \textbf{82.32} & 61.52 & 60.22 & 62.23 & \textbf{62.52} & \textbf{77.44} & 75.20 & 76.16 & 66.56 \\
\multicolumn{1}{c|}{GPT-4o} & 84.12 & 87.83 & \textbf{87.94} & 87.65 & 70.27 & \textbf{71.99} & 71.93 & 71.20 & 79.52 & 75.20 & 73.76 & \textbf{84.80} \\
\hline
\multicolumn{1}{c|}{Gemini 2.0 Flash Thinking} & 87.25 & 86.52 & 86.90 & \textbf{88.99} & \textbf{74.46} & 73.72 & 71.03 & 74.16 & 80.00 & 81.12 & 81.76 & \textbf{86.88} \\
\multicolumn{1}{c|}{GPT o3 mini (low)} & 85.91 & 85.04 & 88.46 & \textbf{88.81} & \textbf{78.29} & 76.39 & 77.02 & 77.26 & 93.60 & 72.16 & 95.04 & \textbf{96.80} \\
\multicolumn{1}{c|}{GPT o3 mini (medium)} & 88.70 & 89.68 & 89.39 & \textbf{89.74} & \textbf{83.60} & 81.14 & 79.27 & 82.18 & \textbf{97.44} & 31.20 & 88.48 & \textbf{97.44} \\
\multicolumn{1}{c|}{GPT o3 mini (high)} & 88.87 & 89.91 & 89.74 & \textbf{91.30} & 85.02 & 83.85 & 81.91 & \textbf{85.08} & \textbf{98.40} & 24.64 & 98.24 & \textbf{98.40} \\
\hline
\end{tabular}
\caption{Accuracy (\%) of different models on BBH-Induct, Evals-Induct, and Shift Cipher in zero-shot settings. The highest overall accuracy is shown in \textbf{bold}. ``Induct'' refers to the INDUCT baseline.}
\label{table:main_result}
\end{table*}

\subsection{Datasets}

We adopt the datasets used in prior state-of-the-art instruction induction methods \cite{chen-etal-2024-induct}, ensuring comparability with existing approaches.

\paragraph{BBH-Induct} This dataset is adapted from the BIG-Bench Hard (BBH) benchmark \cite{Suzgun2022ChallengingBT}, a suite of tasks that current LLMs struggle to match average human performance. It comprises 23 tasks totaling 5,161 instances, covering diverse reasoning challenges including logical deduction, spatial reasoning, and language understanding.

\paragraph{Evals-Induct} Derived from the OpenAI Evals project, this dataset features 24 challenging tasks that are beyond the capabilities of state-of-the-art models. Each task contains approximately 91 to 378 examples, totaling 3,683 instances.

\paragraph{Shift Cipher Task} We additionally include the Shift Cipher Task \cite{prabhakar-etal-2024-deciphering} to assess the effectiveness of our approach on symbolic reasoning tasks. This task involves decoding text encrypted with a shift cipher, where each letter is replaced by another a fixed number of positions earlier in the alphabet. The dataset includes cases with shifts ranging from \( k=1 \) to \( 25 \), requiring models to generalize across different rotation values. Each example consists of a seven-letter word encoded with a shift cipher. Following \citet{prabhakar-etal-2024-deciphering}, we use words from the highest-probability bin (Bin~1), where words are ranked by GPT-2 log probability and may include non-standard forms that are frequent in pre-training corpora but not necessarily valid English words. For instance, a case with $\mathit{shift\_level} = 3$, also known as ROT-3, requires decoding ``fkrrvhg'' back to ``choosed'' by shifting each letter 3 positions backward: \texttt{f} → \texttt{c}, \texttt{k} → \texttt{h}, \texttt{r} → \texttt{o}, and so on.

Further details on all datasets are provided in Table~\ref{tab:task_subtask_shortphrase} to \ref{tab:SC_task_mapping} in the Appendix.

\subsection{Baselines}

\paragraph{ZCoT} 
We apply Zero-shot Chain-of-Thought \cite{Kojima2022LargeLM} using Short Phrase Instructions, where the model receives only a short phrase as guidance for reasoning.

\paragraph{SCoT} 
Automatic Strategic Chain-of-Thought, proposed by \citet{Wang2024StrategicCG}, is a state-of-the-art Chain-of-Thought (CoT) baseline that prompts the model to first identify an effective problem-solving strategy before proceeding with step-by-step reasoning. In our setting, we provide only Short Phrase Instructions as the task description.

\paragraph{INDUCT} 
The instruction induction stage of \citet{chen-etal-2024-induct} serves as the state-of-the-art instruction induction baseline, which applies LLMs to infer task instructions from given examples. In our setting, we provide the model with Short Phrase and Questions (without Answers) to induce task instructions.

ZCoT and SCoT are \textit{instance-level} methods that solve each question independently using only the Short Phrase as the task description. INDUCT and our \textsc{Strategy-Induct} are \textit{task-level} methods that additionally leverage a small set of example questions to induce a reusable task instruction prior to inference. At inference time, all methods receive the same input: one instruction and one new question. The distinction is that instance-level methods use the Short Phrase directly as the instruction, whereas task-level methods replace it with an induced instruction derived from the example questions.

\subsection{Other Details}

All experiments follow the BBH-Induct and Evals-Induct settings, including a Short Phrase for every task or subtask as its task description. The specific phrases for each dataset and subtask are listed in Appendix Tables~\ref{tab:task_subtask_shortphrase}--\ref{tab:SC_task_mapping}.

\paragraph{N Setting}
We set \( N = 3 \), so \textsc{Strategy-Induct} generates a strategy for each of three questions and induces instructions from the resulting pairs.

\paragraph{Experimental Configurations}
We utilize API services from five LLM providers: OpenAI, Google, Mistral AI, SambaNova, and Together AI. Model details are in Appendix~\ref{sec:exp_config}. For cost considerations, we follow the sampling strategy of \citet{chen-etal-2024-induct}, randomly sampling 25 examples per task and setting temperature to 0 for deterministic outputs.

\begin{figure*}[ht!]
    \centering
    \protect\includegraphics[width=\linewidth]{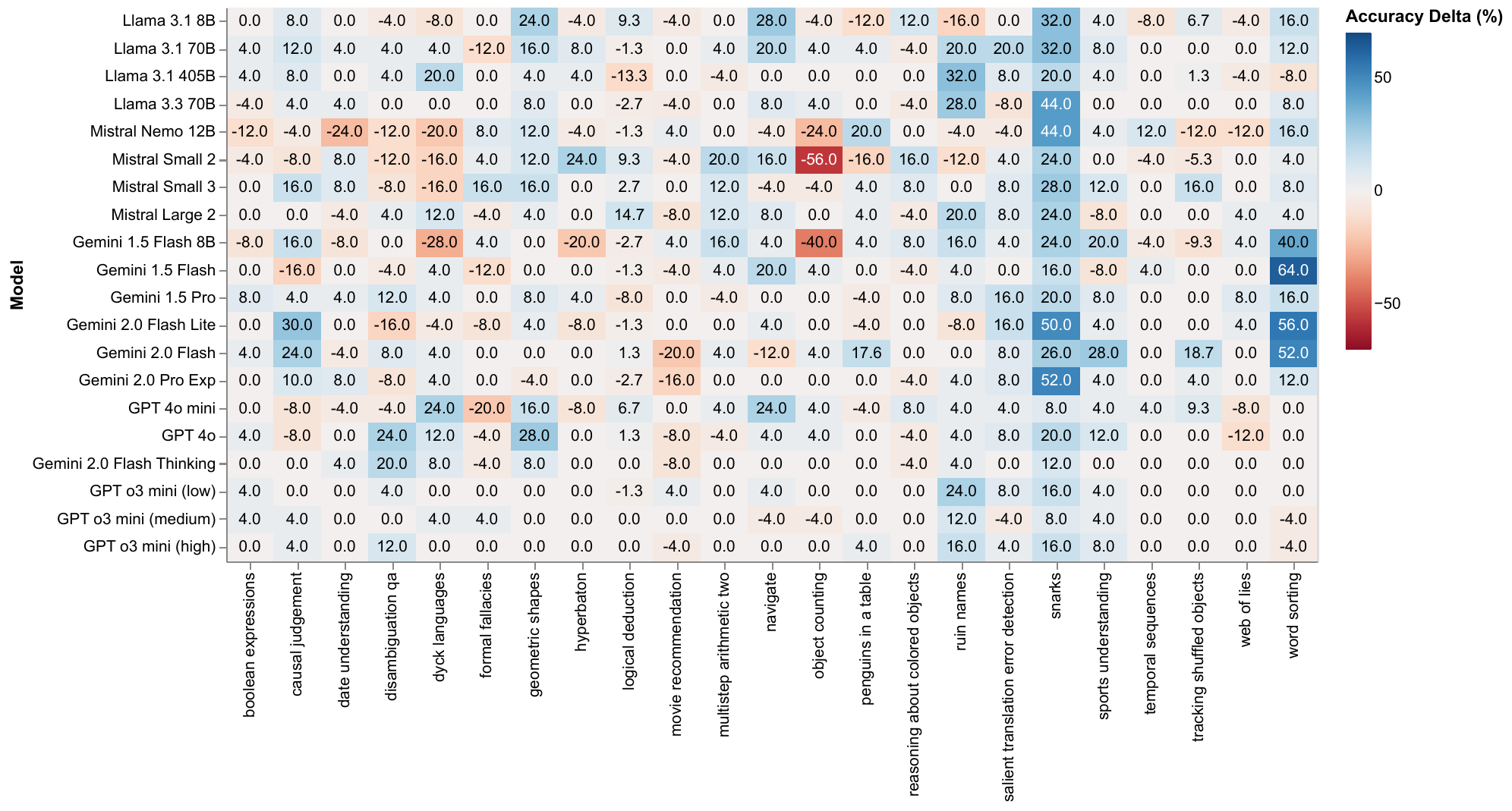}
    \caption{Accuracy difference heatmap for the 23 tasks from BBH-Induct across 18 models. The accuracy delta is computed between \textsc{Strategy-Induct} and the baseline method ZCoT (\textcolor{blue}{blue}/\textcolor{red}{red} indicates our method wins/loses).}
    \label{fig:BBH_head_to_head_heatmap}
\end{figure*}

\begin{figure}[ht!]
    \centering
    \includegraphics[width=0.96\linewidth]{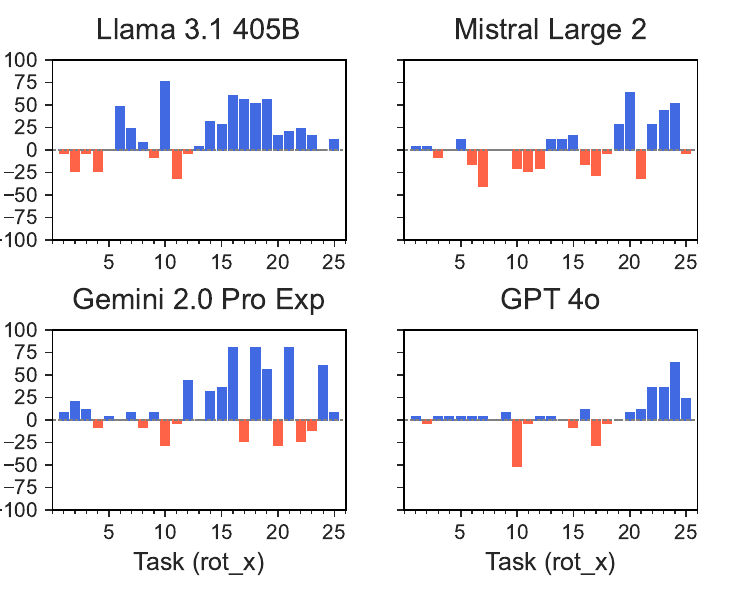}
    \caption{The head-to-head comparison of the largest models within each LLM series on the 25 subtasks from Shift Cipher. The accuracy delta indicates the accuracy difference between \textsc{Strategy-Induct} and ZCoT (\textcolor{blue}{blue}/\textcolor{red}{red} indicates our method wins/loses).}
    \label{fig:SC_head_to_head_LLM}
\end{figure}

\begin{figure}[ht!]
    \centering
    \includegraphics[width=0.96\linewidth]{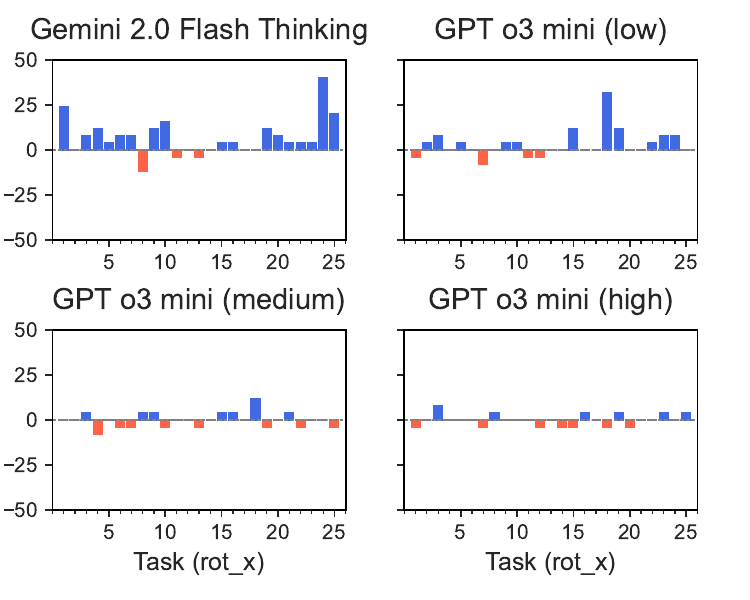}
    \caption{The head-to-head comparison of LRMs on the 25 subtasks from Shift Cipher. The accuracy delta represents the accuracy difference between \textsc{Strategy-Induct} and ZCoT (\textcolor{blue}{blue}/\textcolor{red}{red} indicates our method wins/loses).}
    \label{fig:SC_head_to_head_LRM}
\end{figure}

\section{Results and Analysis}

\subsection{Quality of Strategy-Induced Instructions}

We evaluate \textsc{Strategy-Induct} on BBH-Induct, Evals-Induct, and the Shift Cipher Task using LLMs and LRMs, measuring the accuracy of LLM responses given the question and generated instructions. To assess effectiveness, we compare it with baselines across 60 settings (20 per dataset). As shown in Table~\ref{table:main_result}, our method outperforms existing methods in most settings. Compared to ZCoT, it achieves a 50-3-7 win-tie-lose record. Compared to INDUCT \cite{chen-etal-2024-induct}, it attains 44-3-13, and compared to SCoT, it reaches 52-0-8.

\paragraph{Performance on LLMs} 
Our results align with previous findings \cite{honovich-etal-2023-instruction, chen-etal-2024-induct}, showing that smaller LLMs such as Llama 3.1 8B, Mistral Nemo 12B, Gemini 1.5 Flash 8B, Gemini 2.0 Flash Lite, and GPT-4o mini struggle with instruction induction, making it difficult for them to derive effective task instructions. However, \textsc{Strategy-Induct} mitigates this issue, achieving a 10-3-2 win-tie-lose record against INDUCT on these small models.
For mid-to-large models, \textsc{Strategy-Induct} is particularly effective. Compared to ZCoT, it yields higher accuracy across all models, except for Mistral Large 2, which underperforms on Evals-Induct, and Mistral Small 2, which produces a tie on Shift Cipher. These results demonstrate the scalability of our method across different model sizes.
Additionally, we observe that SCoT struggles with maintaining output format consistency in the Shift Cipher Task, leading to performance degradation. This issue is particularly evident in GPT o3 mini, where incorrect formatting negatively impacts accuracy.

\paragraph{Performance on LRMs}  
A similar trend is observed in LRMs, where stronger models tend to benefit more from our approach. For instance, the win-tie-lose record for GPT o3 mini improves from 2-0-1 (\texttt{low} reasoning effort) to 2-1-0 (\texttt{high} reasoning effort).
Overall, \textsc{Strategy-Induct} improves LLM and LRM performance across diverse tasks, outperforming prior state-of-the-art methods.

\subsection{Breakdown of BBH-Induct}
\label{sec:BBH_breakdown}

We analyze the 23 tasks in BBH-Induct to examine how \textsc{Strategy-Induct} impacts their performance. Figure~\ref{fig:BBH_head_to_head_heatmap} shows accuracy differences across 18 models, comparing \textsc{Strategy-Induct} with the baseline ZCoT.

As model size increases, negative changes (red regions) decrease while positive improvements (blue regions) become more widespread. This trend is consistent with the findings of \citet{chen-etal-2024-induct}. Some exceptions occur in simpler subtasks (e.g., boolean expressions), where strong models such as GPT-4o, Llama 3.1 405B, and LRMs already achieve near-perfect accuracy, leaving little room for improvement.

For traditional LLMs, \textsc{Strategy-Induct} improves performance on semantic understanding tasks, notably \textit{snarks}, \textit{ruin names}, and \textit{sports understanding}. It also improves performance in ranking-related tasks (e.g., \textit{word sorting}), particularly for smaller models. Additionally, tasks involving geometric or spatial reasoning (e.g., \textit{geometric shapes and navigate}) also benefit.

For LRMs, \textsc{Strategy-Induct} improves performance on semantic understanding tasks, such as \textit{disambiguation QA}, \textit{snarks}, \textit{ruin names}, and \textit{sports understanding}. This aligns with the design of LRMs, which primarily focus on mathematical, logical, and programming tasks rather than language understanding.

Notably, several subtasks where \textsc{Strategy-Induct} shows strong gains—\textit{salient translation error detection}, \textit{snarks}, and \textit{sports understanding}—are categorized as knowledge-based rather than reasoning-oriented by \citet{Suzgun2022ChallengingBT}. Our framework yields accuracy improvements of 8 to 60 percentage points on \textit{snarks} and win-tie-lose rates of 15-3-2 and 14-3-3 against ZCoT for \textit{sports understanding} and \textit{salient translation error detection}, respectively. Further analysis is provided in Appendix~\ref{appendix:nonreasoning} and Appendix~\ref{MMLU-PRO_Result}.

We also observed errors in certain smaller models on specific subtasks. For instance, Mistral Nemo, Mistral Small, and Gemini 1.5 Flash 8B tend to classify objects instead of counting them in the \textit{object counting} task, leading to misinterpretations and higher error rates.

\subsection{Breakdown of the Shift Cipher Task}

To better understand the effects of \textsc{Strategy-Induct}, we analyze its performance across different shift values in the Shift Cipher task.
Prior work by \citet{prabhakar-etal-2024-deciphering} showed that even state-of-the-art LLMs, such as GPT-4 and Llama 3.1 405B, perform well on commonly seen shift values like ROT-1, ROT-3 (Caesar cipher), and ROT-13. However, accuracy decreases for less frequent shifts (rot-$k$, where $k \neq 1, 3, 13$).
Following the analytical approach of \citet{prabhakar-etal-2024-deciphering}, we select the largest model from each LLM series as a representative and include all available LRM models to ensure a comprehensive evaluation across different reasoning capabilities.

As shown in Figure~\ref{fig:SC_head_to_head_LLM}, our method achieves greater improvements on less common shift values. A possible explanation is that \textsc{Strategy-Induct} induces prompts that explicitly guide LLMs to handle wrap-around effects in letter positions, thereby reducing errors.
We also observe higher performance fluctuations for ROT-10. For instance, Llama 3.1 405B shows noticeable accuracy gains, whereas GPT-4o exhibits a drop. Our analysis suggests that for ROT-10, LLMs tend to be more sensitive to prompt variations. Specifically, GPT-4o frequently misaligns the positions of 's' and 't', the 19th and 20th letters of the alphabet, respectively, leading to errors. 

Overall, \textsc{Strategy-Induct} enhances LLM performance on the Shift Cipher task (Figure~\ref{fig:SC_head_to_head_LLM}) while also improving reasoning in weaker LRMs (Figure~\ref{fig:SC_head_to_head_LRM}), such as Gemini 2.0 Flash Thinking and GPT o3 mini (low).

\section{Analysis and Discussion}

\subsection{Effect of the Number of Example Questions}

To analyze the impact of the number of example questions ($N$), we conduct experiments with $N=1, 3, 5$ on BBH-Induct. As shown in Table~\ref{table:n_question}, performance generally improves from $N=1$ to $N=3$, with $N=3$ achieving the highest accuracy for most models (19-0-1 win-tie-lose against $N=1$). However, increasing $N$ to 5 does not consistently yield further gains.

A likely explanation is that each example question is paired with a multi-step strategy, so larger $N$ substantially increases the length and complexity of the induction prompt. While $N=3$ offers sufficient diversity for effective induction, $N=5$ may exceed the capacity of some models to synthesize a concise instruction from the additional context.

\begin{table}[ht!]
\small
\centering
\begin{tabular}{c|ccc}
\hline
Model & N=1 & N=3 & N=5 \\ \hline
\multicolumn{1}{c|}{Llama 3.1 8B} & 64.35 & \textbf{65.33} & 61.74 \\
\multicolumn{1}{c|}{Llama 3.1 70B} & 87.54 & 88.99 & \textbf{89.97} \\
\multicolumn{1}{c|}{Llama 3.1 405B} & 88.12 & 88.52 & \textbf{89.10} \\
\multicolumn{1}{c|}{Llama 3.3 70B} & 86.72 & 87.30 & \textbf{87.77} \\
\hline
\multicolumn{1}{c|}{Mistral Nemo 12B} & 51.71 & 55.71 & \textbf{59.13} \\
\multicolumn{1}{c|}{Mistral Small 2} & 66.09 & 70.32 & \textbf{70.43} \\
\multicolumn{1}{c|}{Mistral Small 3} & 75.83 & \textbf{80.41} & 79.59 \\
\multicolumn{1}{c|}{Mistral Large 2} & 84.87 & \textbf{85.97} & 84.58 \\
\hline
\multicolumn{1}{c|}{Gemini 1.5 Flash 8B} & 67.07 & \textbf{69.33} & 68.06 \\
\multicolumn{1}{c|}{Gemini 1.5 Flash} & 80.46 & \textbf{81.80} & \textbf{81.80} \\
\multicolumn{1}{c|}{Gemini 1.5 Pro} & 83.65 & \textbf{85.39} & 84.35 \\
\multicolumn{1}{c|}{Gemini 2.0 Flash Lite} & 81.39 & 81.80 & \textbf{85.08} \\
\multicolumn{1}{c|}{Gemini 2.0 Flash} & 83.10 & 85.45 & \textbf{85.60} \\
\multicolumn{1}{c|}{Gemini 2.0 Pro Exp} & 86.09 & \textbf{86.72} & 85.74 \\
\hline
\multicolumn{1}{c|}{GPT-4o mini} & \textbf{83.65} & 82.32 & 79.30 \\
\multicolumn{1}{c|}{GPT-4o} & 86.67 & \textbf{87.65} & 86.84 \\
\hline
\multicolumn{1}{c|}{Gemini 2.0 Flash Thinking} & 86.87 & \textbf{88.99} & 88.75 \\
\multicolumn{1}{c|}{GPT o3 mini (low)} & 87.83 & \textbf{88.81} & 87.59 \\
\multicolumn{1}{c|}{GPT o3 mini (medium)} & 89.57 & \textbf{89.74} & 88.70 \\
\multicolumn{1}{c|}{GPT o3 mini (high)} & 90.43 & \textbf{91.30} & 90.61 \\
\hline
\end{tabular}
\caption{Effect of the number of example questions $N$ on \textsc{Strategy-Induct} performance across different models on BBH-Induct.}
\label{table:n_question}
\end{table}

\begin{figure*}[ht!]
    \centering
    \protect\includegraphics[width=1.00\linewidth]{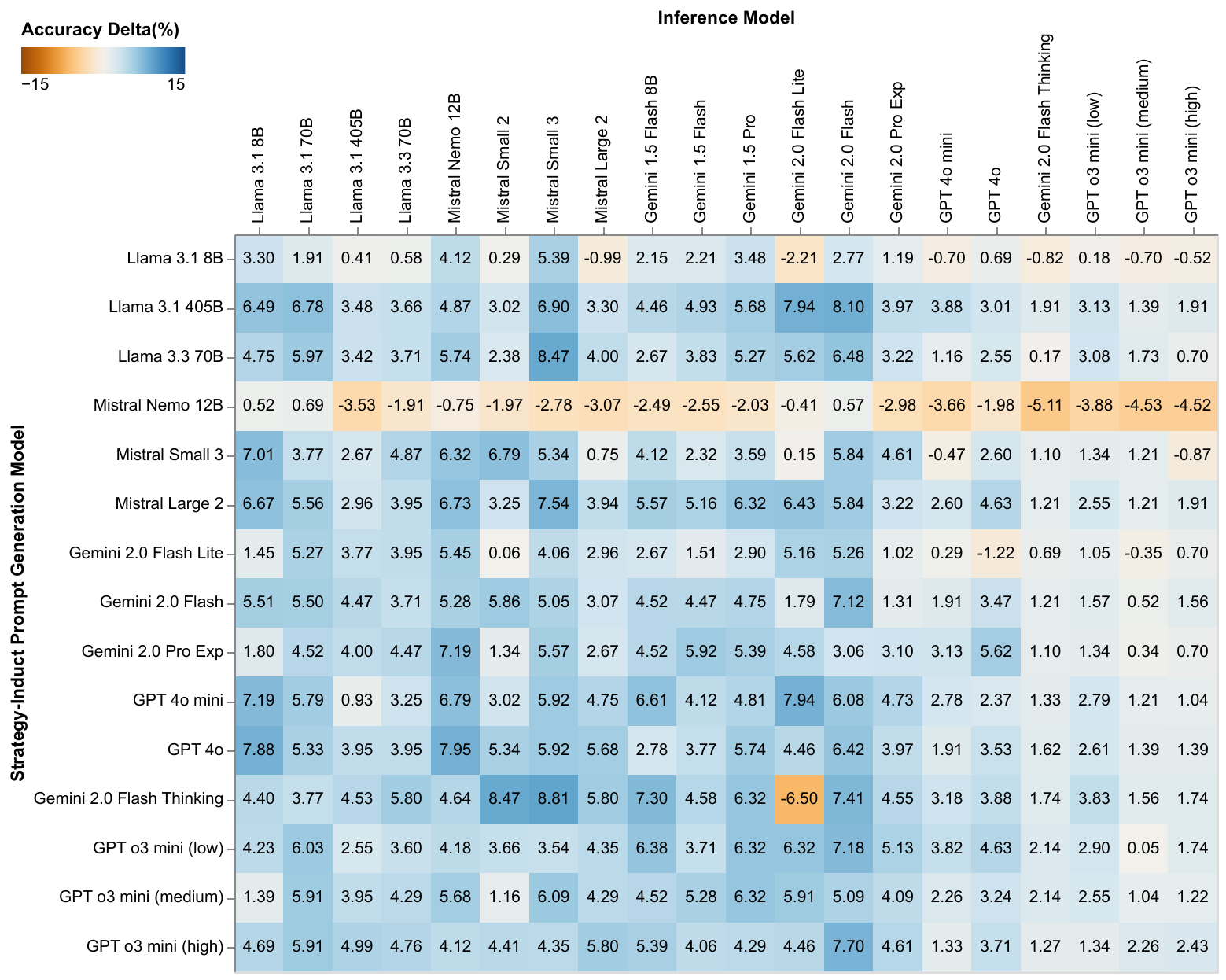}
    \caption{Cross-model generalization analysis on BBH-Induct, where the prompt induction and inference models are different. Accuracy is shown as a percentage, with darker \textcolor{blue}{blue} indicating greater improvement and darker \textcolor{orange}{orange} indicating performance decline.}
    \label{fig:BBH_cross_model}
\end{figure*}

\subsection{Cross-Model Generalization}

To evaluate the cross-model capability of our approach, we select recent small, medium, and large models from each LLM series and assess whether the prompts they generate enhance the performance of various inference models.  
We analyze this generalization from four perspectives:  
(1) Effect of Model Scale on Prompt Quality,  
(2) Generalization of Prompts Across LLM Series,  
(3) Impact of Prompts on Different Model Scales, and  
(4) Cooperation Between LLMs and LRMs. Figure~\ref{fig:BBH_cross_model} presents our experimental results, which we analyze below.  

\subsubsection{Effect of Model Scale on Prompt Quality}
Smaller models, such as Llama 3.1 8B and Mistral Nemo 12B, tend to generate less effective prompts, sometimes even reducing the performance of downstream models.  
In contrast, prompts from larger models often lead to widespread performance improvements. For example, those generated by Llama 3.1 405B and Mistral Large 2 contribute to notable improvements.  
Additionally, we observe that although Gemini 2.0 Pro Exp is larger than Gemini 2.0 Flash, their performance on BBH-Induct is similar (86.72\% vs. 85.45\%), resulting in prompts of comparable quality. This suggests that prompt quality correlates more with task performance than with model size alone.

\subsubsection{Generalization of Prompts Across LLM Series}
Our approach demonstrates strong cross-series generalization. The most capable models in each series (Llama 3.1 405B, Mistral Large 2, Gemini 2.0 Pro Exp, and GPT-4o) generate prompts that improve the performance of models from other series.  

We observe that applying prompts across different models can sometimes yield higher performance gains.
For example, when GPT-4o uses its own induced prompts on BBH-Induct, it achieves a 3.53\% improvement over the baseline.
 
However, when using prompts generated by Mistral Large 2 and Gemini 2.0 Pro Exp, the improvement increases to 4.63\% and 5.62\%, respectively.
This suggests that prompts generated by different models may introduce alternative reasoning strategies, helping the original model refine its outputs, a pattern consistent across other model families.

\subsubsection{Impact of Prompts on Different Model Scales}
Smaller models benefit more from detailed prompts, as they rely more on explicit guidance.  
For example, in the Llama 3.1 series, when using prompts from GPT-4o, the performance gains for Llama 3.1 8B, 70B, and 405B are 7.88\%, 5.33\%, and 3.95\%, respectively.  
This pattern suggests that high-quality prompts provide greater improvements for smaller models.

\begin{figure*}[t!]
\scriptsize
\begin{tcolorbox}[title={Case Study: Snarks Task (BBH)}]

\textbf{Input Question and Generated Strategies:}

\vspace{2pt}
\textbf{Input Question 1:}\\
\quad (A) Think of the poor man! What will he do without his job?\\
\quad (B) Think of the poor man! What will he do without his yacht?\\
\textbf{Generated Strategy 1:}\\
\quad Step 1: Analyze option (A) [...] Consider the common understanding of job loss as a serious issue.\\
\quad Step 2: Analyze option (B) [...] Consider owning a yacht as a luxury, and the potential for sarcasm when expressing concern about someone losing a luxury item. [...]

\vspace{3pt}
\textbf{Input Question 2:}\\
\quad (A) What a poor guy, years of work and he only got a \$50,000 scholarship\\
\quad (B) What a poor guy, years of work and he only got a \$50 scholarship\\
\textbf{Generated Strategy 2:}\\
\quad Step 1: Analyze option (A) for potential sarcasm. Consider the implied meaning of ``poor guy'' in relation to receiving a \$50,000 scholarship. Evaluate if the statement conveys the opposite of its literal meaning.\\
\quad Step 2: Analyze option (B) [...] \quad Step 3: Compare the two options (A) and (B). [...]

\vspace{3pt}
\textbf{Input Question 3:}\\
\quad (A) Have you tried not being poor? It is much easier\\
\quad (B) Have you tried not being rude? It is much easier\\
\textbf{Generated Strategy 3:}\\
\quad Step 1: Analyze option A for sarcastic intent. Consider the inherent difficulty of overcoming poverty and whether the statement implies the opposite.\\
\quad Step 2: Analyze option B for sarcastic intent. Consider the ease of being polite and whether the statement implies the opposite. [...]

\tcblower

\textbf{Final Induced Instruction (\textsc{Strategy-Induct}):}\\[2pt]
Step 1: Analyze Sentence (A): Carefully examine sentence (A) for potential indicators of sarcasm. [...] Look for elements such as:\\
\quad -- Irony: A statement where the intended meaning is the opposite of the literal meaning.\\
\quad -- Hyperbole: Exaggeration for emphasis or effect.\\
\quad -- Satire: The use of humor, irony, exaggeration, or ridicule... [...]\\
Step 2: Analyze Sentence (B): Repeat the analysis process from Step 1 for sentence (B).\\
Step 3: Compare Sentences: Compare the two sentences based on the likelihood of sarcastic intent. [...]\\
Step 4: Select the More Sarcastic Sentence: Determine which sentence (A or B) is more likely to be sarcastic based on your analysis.

\vspace{4pt}
\textbf{Result (Gemini 2.0 Flash):} SCoT: 0.40 $\rightarrow$ \textsc{Strategy-Induct}: \textbf{0.76}
\end{tcolorbox}
\caption{Case study on the \textit{snarks} task (Gemini 2.0 Flash). Each instance-level strategy captures one sarcasm cue; the induced instruction unifies them into a comprehensive reasoning framework. Full details in Appendix.}
\label{fig:case-study-main}
\end{figure*}

\subsubsection{Cooperation Between LLMs and LRMs}
LRMs sometimes gain less from their self-generated prompts compared to those from stronger LLMs.  
For instance, prompts generated by Llama 3.1 405B, when used by GPT o3 mini (low and medium), lead to greater performance gains compared to prompts generated by GPT o3 mini itself.

Meanwhile, some LRMs generate prompts that sometimes perform on par with, or even surpass, those from larger LLMs.  
For example, prompts from Gemini 2.0 Flash Thinking significantly boost Mistral Small 2 and Mistral Small 3, yielding gains of 8.47\% and 8.81\%, respectively.

However, while Gemini 2.0 Flash Thinking's prompts improve most models, they negatively impact Gemini 2.0 Flash Lite.  
A possible explanation is that Flash Thinking tends to generate more structured, multi-step strategies, which, while beneficial for many models, may be overly complex for Flash Lite.  
For instance, in tasks like \textit{Tracking Shuffled Objects}, where precise step tracking is required, Flash Lite may struggle to follow these detailed instructions, leading to errors.  

These findings suggest that LLMs and LRMs may complement each other, with LLMs potentially enhancing task instruction quality for LRMs, while certain LRMs may generate prompts that improve other models.

\subsection{Case Study: From Instance Strategies to Task Instructions}

Figure~\ref{fig:case-study-main} illustrates how \textsc{Strategy-Induct} abstracts instance-level insights into a reusable task instruction on the \textit{snarks} task. Each individual strategy captures only one sarcasm cue (e.g., irony or hyperbole), whereas the induced instruction consolidates them into a comprehensive framework covering irony, hyperbole, satire, and other indicators. This generalization leads to a substantial accuracy gain on Gemini 2.0 Flash, from 0.40 (SCoT) to 0.76 (\textsc{Strategy-Induct}).

\section{Related Work}

\paragraph{Inductive Reasoning in LLMs}
Inductive reasoning, a fundamental cognitive skill, has been widely examined in LLMs. Prior studies show that LLMs can infer patterns and rules from limited observations \cite{Liu2024AnIL, Yu2023NaturalLR, Li2024MIRAGEEA}. \citet{he2024ideaenhancingrulelearning} found that while LLMs exhibit inherent inductive capabilities, interactive environments enhance their effectiveness. 
\citet{Yang2022LanguageMA} investigated rule induction from factual data, while \citet{Zhu2023LargeLM} explored generalization from examples. \citet{10581497} further incorporated inductive reasoning into instruction fine-tuning, and \citet{chen-etal-2024-temporal} applied it to enhance temporal reasoning. These studies highlight the potential of leveraging induction to improve LLMs' reasoning abilities.

\paragraph{Task-Level Instruction Generation}

Instruction induction, where LLMs generate task instructions from input-output examples, was introduced by \citet{honovich-etal-2023-instruction}. Subsequent studies have explored various techniques, including ranking candidate instructions \cite{Zhou2022LargeLM, zhang-etal-2023-auto}, iterative refinement using demonstrations \cite{Sun2023AutoHintAP, Chen2023InstructZeroEI}, and dynamic reasoning modules \cite{aswani-etal-2024-auto}. Additionally, evolutionary optimization has also been applied to improve instruction generation \cite{Xu2022GPSGP, Fernando2023PromptbreederSS, DBLP:conf/iclr/Guo0GLS0L0Y24}. However, these methods often require substantial computational resources and are typically optimized for a single model, limiting their generalizability.

More efficient alternatives focus on generating a single instruction with minimal overhead. \citet{Zhou2024SelfDiscoverLL} constructed reasoning structures from predefined modules, while \citet{chen-etal-2024-induct} proposed an approach that generates a single instruction with chain-of-thought (CoT) demonstrations to improve efficiency. However, the approach of \citet{chen-etal-2024-induct} still relies on labeled input-output pairs for the induction process. Our proposed \textsc{Strategy-Induct} framework follows a similar efficiency-driven approach but addresses a fundamentally different setting. Unlike prior work that requires labeled input-output pairs for induction, \textsc{Strategy-Induct} derives task instructions from a small set of example questions alone, using strategy generation as the key mechanism to compensate for the absence of labeled answers. 

Additionally, our method is model-agnostic, demonstrating strong cross-model generalization. Beyond text, this low-cost induction paradigm has also been extended to multimodal settings, including visual question answering \cite{10.1145/3746252.3760963} and audio understanding \cite{chen-etal-2026-task}.

\paragraph{Reasoning Methods} 
Chain-of-Thought (CoT) prompting improves LLMs' problem-solving by generating intermediate reasoning steps \cite{Wei2022ChainOT,Kojima2022LargeLM,Nye2021ShowYW}. To enhance CoT effectiveness, prior work has explored diverse example selection \cite{Zhang2022AutomaticCO} and structured reasoning strategies, including planning before answering \cite{Wang2023PlanandSolvePI} and formulating strategies first \cite{Wang2024StrategicCG}. \citet{chen-etal-2025-diverge} further proposed generating multiple diverse rationales per question and inducing them into a unified plan, improving zero-shot reasoning without example questions. 

However, this approach operates at the instance level, requiring the full multi-rationale induction process for every new question, which incurs higher inference cost. These methods all require per-instance reasoning, generating a reasoning path for each query. In contrast, \textsc{Strategy-Induct} operates at the task level, deriving a reusable task instruction from a small set of example questions without answer labels, amortizing the induction cost across all questions within the same task.

\section{Conclusion}
We propose \textsc{Strategy-Induct}, a framework that derives task-level instructions solely from example questions without requiring labeled answers. By generating reasoning strategies for each question and inducing a unified instruction from these strategy-question pairs, our approach enables effective instruction generation in question-only settings. By leveraging strategy generation as a substitute for labeled outputs, \textsc{Strategy-Induct} broadens the applicability of instruction induction to scenarios where annotated data is unavailable. Experiments across multiple datasets and model families show that \textsc{Strategy-Induct} outperforms existing methods across various LLMs and LRMs, demonstrating strong adaptability across model scales. Our findings further suggest that combining LLMs and LRMs for task instruction generation and inference may yield additional performance gains. The framework also exhibits robust cross-model generalization, highlighting its broad applicability and potential for future extensions.

\section{Limitations}

\paragraph{Reliance on Instruction-Following Models}  
Our approach relies on sufficiently strong instruction following models to generate responses in the desired format and to induce instructions that facilitate the extraction of effective strategies. While modern instruction following models, such as Llama 3.1 8B, Mistral NeMo 12B, Gemini 1.5 Flash 8B, and GPT-4o mini, are capable of handling complex instructions, earlier models, such as Llama 1 and Llama 2, may lack the necessary capability to adhere to structured formats.

\paragraph{Generalization Across Models and Datasets}  
Our approach has been evaluated on multiple datasets and across 18 different models (with 20 settings), covering a diverse range of architectures and training paradigms. While our results demonstrate strong generalization across these settings, its applicability to models and datasets beyond our experimental scope remains to be further explored.

\paragraph{Dependence on API-Based Models}  
Due to hardware limitations, our experiments rely on API-based access to open-source models provided by services such as Mistral AI, SambaNova, and Together AI. Since we do not have full control over the internal configurations of these services, there may be variations in model behavior across different API providers.

\paragraph{Task Scope}
Our framework is designed for task-level instruction induction, where a single reusable instruction is derived for an entire category of problems within established task definitions. It excels at handling repetitive problems that share a common structure, such as those in BBH-Induct, Evals-Induct, and the Shift Cipher Task. However, real-world tasks can be diverse, open-set, or combinations of multiple tasks, where preparing task descriptions and collecting representative example questions may be more challenging. Extending our approach to such open-ended scenarios remains an area for future work.

\section*{Acknowledgements}
This work was supported by National Science and Technology Council, Taiwan, under grant NSTC 114-2221-E-002 -070 -MY3, and by Ministry of Education (MOE), Taiwan, under grant NTU-114L900901.

\bibliography{custom}

\appendix

\clearpage

\section{Prompt Templates}

We use three types of prompts: Strategy Generation, Induction (including both Instruction Induction and Strategy Induction), and Inference.

To structure the reasoning process effectively, we adopt \texttt{<deduction>} to encapsulate intermediate reasoning steps and \texttt{<final\_answer>} to indicate the conclusive response.
During evaluation, we directly extract the content within \texttt{<final\_answer>} for answer matching.
For detailed prompt structures, please refer to \figurename~\ref{fig:Strategy_Design_Template} to \ref{fig:Induct_STRATEGY_INDUCT_Template}.

\section{Detailed Experimental Configurations}
\label{sec:exp_config}

\subsection{Models Used in Experiments}

We evaluate our framework using 18 models, including four series of large language models (LLMs) and reasoning models (LRMs), covering both open-source and proprietary systems.

\paragraph{Llama models} 
Llama 3.1 8B, Llama 3.1 70B, Llama 3.1 405B, and Llama 3.3 70B~\cite{Dubey2024TheL3}.

\paragraph{Mistral models} 
Mistral Nemo 12B, Mistral Small 2, Mistral Small 3, and Mistral Large 2~\cite{mistral2024nemo, mistral2024small2, mistral2024large2_1,mistral2025small3}.

\paragraph{Gemini models} 
Gemini 1.5 Flash 8B, Gemini 1.5 Flash, Gemini 1.5 Pro, Gemini 2.0 Flash Lite Preview, Gemini 2.0 Flash, and Gemini 2.0 Pro Exp~\cite{Reid2024Gemini1U, google2025gemini2}.

\paragraph{GPT models}
GPT-4o mini and GPT-4o~\cite{openai2024gpt4omini, openai2024gpt4o}.

\paragraph{LRMs}  
Gemini 2.0 Flash Thinking Exp \cite{google2025gemini2}  and GPT o3 mini \cite{openai_2025_o3_mini} with \texttt{low}, \texttt{medium}, and \texttt{high} reasoning effort settings.

\subsection{Other Details}

\paragraph{Model Version}
Table~\ref{table:model_detail} lists the exact model versions used in our experiments, following the naming conventions used in the API.

\begin{table}[h!]
\scriptsize
\begin{tabular}{c|c}
\hline
\textbf{Model} & \textbf{\texttt{Model Version}} \\
\hline
Llama 3.1 8B & \texttt{Meta-Llama-3.1-8B-Instruct} \\
\hline
Llama 3.1 70B & \texttt{Meta-Llama-3.1-70B-Instruct} \\
\hline
Llama 3.1 405B & \texttt{Meta-Llama-3.1-405B-Instruct} \\
\hline
Llama 3.3 70B & \texttt{Llama-3.3-70B-Instruct-Turbo} \\
\hline
Mistral Nemo 12B & \texttt{open-mistral-nemo-2407} \\
\hline
Mistral Small 2 & \texttt{mistral-small-2409} \\
\hline
Mistral Small 3 & \texttt{mistral-small-2501} \\
\hline
Mistral Large 2 & \texttt{mistral-large-2411} \\
\hline
Gemini 1.5 Flash 8B & \texttt{gemini-1.5-flash-8b-001} \\
\hline
Gemini 1.5 Flash & \texttt{gemini-1.5-flash-002} \\
\hline
Gemini 1.5 Pro & \texttt{gemini-1.5-pro-002} \\
\hline
\shortstack{Gemini 2.0 Flash\\Lite Preview} & \texttt{gemini-2.0-flash-lite-preview-02-05} \\
\hline
Gemini 2.0 Flash & \texttt{gemini-2.0-flash-001} \\
\hline
Gemini 2.0 Pro Exp & \texttt{gemini-2.0-pro-exp-02-05} \\
\hline
GPT-4o mini & \texttt{gpt-4o-mini-2024-07-18} \\
\hline
GPT-4o & \texttt{gpt-4o-2024-11-20} \\
\hline
\shortstack{Gemini 2.0 Flash\\Thinking Exp} & \texttt{gemini-2.0-flash-thinking-exp-01-21} \\
\hline
GPT o3 mini & \texttt{o3-mini-2025-01-31} \\
\hline
\end{tabular}
\caption{Correspondence between model names and detailed versions.}
\label{table:model_detail}
\end{table}

\begin{table}[h!]
\scriptsize
\begin{tabular}{crc}
\hline
\textbf{Model} & \textbf{Cost (USD)} & \textbf{API Provider} \\ \hline
Llama 3.1 8B & 3 & SambaNova \\
Llama 3.1 70B & 20 & SambaNova \\
Llama 3.1 405B & 168 & SambaNova \\
Llama 3.3 70B & 21 & Together AI \\
Mistral Nemo 12B & 4 & Mistral AI \\
Mistral Small 2 & 9 & Mistral AI \\
Mistral Small 3 & 4 & Mistral AI \\
Mistral Large 2 & 86 & Mistral AI \\
Gemini 1.5 Flash 8B & 2 & Google \\
Gemini 1.5 Flash & 4 & Google \\
Gemini 1.5 Pro & 66 & Google \\
Gemini 2.0 Flash Lite Preview & 4 & Google \\
Gemini 2.0 Flash & 5 & Google \\
Gemini 2.0 Pro Exp & - & Google \\
GPT-4o mini & 8 & OpenAI \\
GPT-4o & 132 & OpenAI \\
Gemini 2.0 Flash Thinking Exp & - & Google \\
GPT o3 mini (low) & 73 & OpenAI \\
GPT o3 mini (medium) & 139 & OpenAI \\
GPT o3 mini (high) & 224 & OpenAI \\
\textbf{Total Cost} & 972 &  \\ \hline
\end{tabular}
\caption{Estimated total costs for each model (Google experimental models are free and are denoted by '-' in the table, thus not included in the cost calculation).}
\label{table:cost}
\end{table}

\paragraph{Safety Settings}
We disable all safety settings for Gemini to prevent Google’s API from refusing to respond. 

\paragraph{Cost Details}

Table~\ref{table:cost} shows the estimated total cost for all experiments, which amounted to approximately \$972 with API services provided by
SambaNova, Together AI, Mistral AI, Google, and OpenAI.

\paragraph{Model Settings}
We set the temperature to 0 and Top-P to 1. The maximum output length is set to 8,192 tokens, if supported by the model.

\section{Use of AI Assistants}  

In this paper, GPT-4o and GPT o3 mini were used for grammar refinement and assistance in code development. However, all outputs from the language models served only as references, and the final writing of the paper and code was reviewed and authored by the researcher.

\section{Detailed Dataset Information}
\label{appendix:dataset_detail}
This appendix provides the Short Phrases used for different tasks across all datasets in our study. We adopt the Short Phrases from \cite{chen-etal-2024-induct}, making minor adjustments for clarity where necessary. The specific Short Phrases for each task are listed in Table~\ref{tab:task_subtask_shortphrase} to Table~\ref{tab:SC_task_mapping}.

\begin{table*}[ht!]
\small
\begin{tabular}{l|lll|lll|lll}
\hline
\multicolumn{1}{c|}{\textbf{Model}} & \multicolumn{3}{c|}{\textbf{Salient.}} & \multicolumn{3}{c|}{\textbf{Snarks}} & \multicolumn{3}{c}{\textbf{Sports}} \\
\multicolumn{1}{c|}{} & \multicolumn{1}{c}{\textbf{ZCoT}} & \multicolumn{1}{c}{\textbf{Induct}} & \multicolumn{1}{c|}{\textbf{Our}} & \multicolumn{1}{c}{\textbf{ZCoT}} & \multicolumn{1}{c}{\textbf{Induct}} & \multicolumn{1}{c|}{\textbf{Our}} & \multicolumn{1}{c}{\textbf{ZCoT}} & \multicolumn{1}{c}{\textbf{Induct}} & \multicolumn{1}{c}{\textbf{Our}} \\ \hline
Llama 3.1 8B & \textbf{56\%} & 48\% & \textbf{56\%} & 32\% & 24\% & \textbf{64\%} & 80\% & 76\% & \textbf{84\%} \\
Llama 3.1 70B & 64\% & 72\% & \textbf{84\%} & 60\% & 88\% & \textbf{92\%} & 84\% & 84\% & \textbf{92\%} \\
Llama 3.1 405B & 68\% & \textbf{88\%} & 76\% & 60\% & \textbf{84\%} & 80\% & 84\% & \textbf{96\%} & 88\% \\
Llama 3.3 70B & \textbf{80\%} & 76\% & 72\% & 48\% & 84\% & \textbf{92\%} & \textbf{88\%} & \textbf{88\%} & \textbf{88\%} \\ \hline
Mistral Nemo 12B & \textbf{56\%} & 52\% & 52\% & 24\% & 72\% & \textbf{84\%} & 80\% & 44\% & \textbf{88\%} \\
Mistral Small 2 & 68\% & 60\% & \textbf{72\%} & 24\% & \textbf{88\%} & 48\% & \textbf{96\%} & \textbf{96\%} & \textbf{96\%} \\
Mistral Small 3 & 64\% & 68\% & \textbf{72\%} & 48\% & 68\% & \textbf{76\%} & 88\% & \textbf{100\%} & \textbf{100\%} \\
Mistral Large 2 & 76\% & 76\% & \textbf{84\%} & 60\% & \textbf{92\%} & 84\% & 92\% & \textbf{96\%} & 84\% \\ \hline
Gemini 1.5 Flash 8B & 56\% & 56\% & \textbf{60\%} & 28\% & \textbf{60\%} & 52\% & 60\% & 60\% & \textbf{80\%} \\
Gemini 1.5 Flash & 64\% & \textbf{76\%} & 64\% & 40\% & \textbf{68\%} & 56\% & \textbf{88\%} & 84\% & 80\% \\
Gemini 1.5 Pro & 64\% & 72\% & \textbf{80\%} & 60\% & 76\% & \textbf{80\%} & 80\% & \textbf{92\%} & 88\% \\
Gemini 2.0 Flash Lite & 64\% & 60\% & \textbf{80\%} & 30\% & 72\% & \textbf{80\%} & 76\% & \textbf{88\%} & 80\% \\
Gemini 2.0 Flash & 64\% & 64\% & \textbf{72\%} & 50\% & \textbf{76\%} & \textbf{76\%} & 56\% & \textbf{92\%} & 84\% \\
Gemini 2.0 Pro Exp & 68\% & 64\% & \textbf{76\%} & 40\% & 88\% & \textbf{92\%} & 88\% & \textbf{92\%} & \textbf{92\%} \\ \hline
GPT 4o mini & 64\% & \textbf{68\%} & \textbf{68\%} & 68\% & \textbf{80\%} & 76\% & 88\% & \textbf{96\%} & 92\% \\
GPT 4o & 68\% & 72\% & \textbf{76\%} & 68\% & 80\% & \textbf{88\%} & 84\% & \textbf{96\%} & \textbf{96\%} \\ \hline
Gemini 2.0 Flash Thinking & 68\% & \textbf{72\%} & 68\% & 76\% & 80\% & \textbf{88\%} & \textbf{88\%} & 84\% & \textbf{88\%} \\
GPT o3 mini (low) & 68\% & 68\% & \textbf{76\%} & 52\% & \textbf{76\%} & 68\% & 92\% & 88\% & \textbf{96\%} \\
GPT o3 mini (medium) & \textbf{76\%} & \textbf{76\%} & 72\% & 72\% & 76\% & \textbf{80\%} & 92\% & \textbf{96\%} & \textbf{96\%} \\
GPT o3 mini (high) & 68\% & \textbf{72\%} & \textbf{72\%} & 76\% & 72\% & \textbf{92\%} & 88\% & \textbf{100\%} & 96\% \\ \hline
\end{tabular}
\caption{Accuracy comparison on three non-reasoning tasks from BBH-Induct (Salient Translation Error Detection, Snarks, and Sports) across 18 models. "Salient." refers to the Salient Translation Error Detection task. Bold numbers indicate the best performance for each model.}
\label{tab:non-reasoning-tasks}
\end{table*}

\section{Analysis of Non-Reasoning Tasks in BBH-Induct}
\label{appendix:nonreasoning}

As discussed in Section~\ref{sec:BBH_breakdown}, although BBH is primarily a reasoning-oriented benchmark, several subtasks—such as \textit{salient translation error detection}, \textit{snarks}, and \textit{sports understanding}—are classified as knowledge-based tasks by \citet{Suzgun2022ChallengingBT}.

Table~\ref{tab:non-reasoning-tasks} presents a direct comparison among Z-CoT, the Induct baseline, and our framework on these subtasks. Our method achieves a win-tie-lose rate of 30-10-20 compared to Induct, and 49-6-5 compared to Z-CoT. These results indicate that strategy induction can be beneficial for a range of non-reasoning tasks, in addition to its strengths on reasoning-oriented benchmarks. 
Further experiments and results on additional non-reasoning tasks are presented in the following section.

\section{Additional Experimental Result}

Our current setting follows Induct-Learn \citep{chen-etal-2024-induct}, where the authors constructed the BBH-Induct dataset by extracting task-relevant context from the original BBH questions, and removing them from the original question. This added context, referred to as the \textbf{Short Phrase}. The following is an example of BBH-Induct:

\begin{tcolorbox}[
  title=Example from Short Phrase Task,  
]
\small
\textbf{Short Phrase:} \\ 
Similar Movie Recommendation \\
\\
\textbf{Question:}  \\
Minority Report, Total Recall, Inside Out, Forrest Gump \\
\\
\textbf{Options:}  \\
(A) Phenomena \quad \\
(B) Lilting \quad  \\
(C) Catwoman \quad  \\
(D) Edge of Tomorrow \\
\end{tcolorbox}

Without the Short Phrase, the question appears as just a list of movie names and options, and the task objective becomes ambiguous. This is a specific characteristic of the BBH-Induct dataset. However, in many real-world settings (even in the original BBH dataset), user queries tend to include the task objective explicitly. In such cases, the Short Phrase becomes optional.

\subsection{MMLU-PRO Result}
\label{MMLU-PRO_Result}

To further investigate this, we conducted additional experiments comparing the performance with and without Short Phrases. We chose the MMLU-PRO dataset for this comparison, as it consists of knowledge-based questions that better reflect non-reasoning tasks, which also allows us to evaluate the generalizability of our framework beyond reasoning-oriented benchmarks. This design allowed us to systematically examine both issues within a single experimental setting.

We first present the performance comparison on the MMLU-PRO dataset with Short Phrases included, to highlight how our framework performs against baselines under this setting.

Due to computational constraints, we sampled 25 examples from each of the 14 sub-tasks in MMLU-PRO, resulting in a total of 350 instances. We evaluated our framework across 15 different models, and compared it against the current state-of-the-art Induct-Learn (Induct) and Zero-Shot Chain of Thought (ZCoT).
The experimental results are presented in Table~\ref{tab:MMLU-PRO-Result}.

\begin{table}[t!]
\footnotesize
\centering
\setlength{\tabcolsep}{4pt}
\begin{tabular}{llll}
\hline
\textbf{Model} & \textbf{ZCoT} & \textbf{Induct} & \textbf{Our} \\ \hline
Llama 3.1 8B & 41.71\% & \textbf{44.00\%} & 42.00\% \\
Llama 3.3 70B & 70.29\% & 69.14\% & \textbf{71.43\%} \\ \hline
Mistral Small 2 & \textbf{49.71\%} & 47.14\% & 46.86\% \\
Mistral Small 3 & 64.86\% & \textbf{67.43\%} & 66.29\% \\
Mistral Large 2 & 68.57\% & 68.57\% & \textbf{68.86\%} \\ \hline
Gemini 1.5 Flash 8B & 52.29\% & 53.71\% & \textbf{54.57\%} \\
Gemini 1.5 Flash & \textbf{66.00\%} & 65.14\% & 65.71\% \\
Gemini 1.5 Pro & 74.86\% & 73.43\% & \textbf{76.29\%} \\
Gemini 2.0 Flash Lite & \textbf{68.29\%} & \textbf{68.29\%} & 67.71\% \\
Gemini 2.0 Flash & 63.14\% & 62.86\% & \textbf{70.29\%} \\
Gemini 2.0 Pro Exp & 82.29\% & \textbf{84.29\%} & 82.86\% \\ \hline
GPT-4o mini & 62.29\% & \textbf{63.14\%} & 60.57\% \\
GPT-4o & 75.14\% & 76.00\% & \textbf{76.57\%} \\ \hline
Gemini 2.0 Flash Thinking & 75.43\% & 75.14\% & \textbf{75.71\%} \\
GPT o3 mini (low) & 76.57\% & 74.00\% & \textbf{77.14\%} \\
GPT o3 mini (medium) & 77.71\% & 77.43\% & \textbf{78.57\%} \\
GPT o3 mini (high) & 79.71\% & 79.14\% & \textbf{80.00\%} \\ \hline
\end{tabular}
\caption{Accuracy (\%) of different models on MMLU-PRO in zero-shot settings.
The highest overall accuracy is shown in bold. “Induct” refers to the INDUCT baseline.}
\label{tab:MMLU-PRO-Result}
\end{table}

\subsection{Effect of Short Phrases}

To further examine the role of Short Phrases, we compare the performance of our framework with and without Short Phrases using the same experimental protocol described above. Table~\ref{fig:with_short_phrase} presents the results for all evaluated models under both settings.

We observe that the performance differences across models are generally minor. This outcome can be attributed to the nature of the MMLU-PRO dataset: the questions are already well-structured and self-explanatory, so the inclusion of an explicit Short Phrase offers limited additional benefit. 

These results indicate that, for benchmarks like MMLU-PRO where questions themselves provide sufficient task context, Short Phrases are optional and do not significantly impact performance.

\begin{table}[t!]
\footnotesize
\centering
\begin{tabular}{lll}
\hline
\textbf{Model} & \textbf{Non SP} & \textbf{SP} \\ \hline
Llama 3.1 8B & \textbf{47.43\%} & 42.00\% \\
Llama 3.3 70B & 70.86\% & \textbf{71.43\%} \\ \hline
Mistral Small 2 & \textbf{49.43\%} & 46.86\% \\
Mistral Small 3 & 66.00\% & \textbf{66.29\%} \\
Mistral Large 2 & \textbf{69.71\%} & 68.86\% \\ \hline
Gemini 1.5 Flash 8B & 54.00\% & \textbf{54.57\%} \\
Gemini 1.5 Flash & \textbf{66.00\%} & 65.71\% \\
Gemini 1.5 Pro & 76.00\% & \textbf{76.29\%} \\
Gemini 2.0 Flash Lite & \textbf{69.71\%} & 67.71\% \\
Gemini 2.0 Flash & 68.29\% & \textbf{70.29\%} \\
Gemini 2.0 Pro Exp & \textbf{84.00\%} & 82.86\% \\ \hline
GPT-4o mini & 60.29\% & \textbf{60.57\%} \\
GPT-4o & 76.00\% & \textbf{76.57\%} \\ \hline
Gemini 2.0 Flash Thinking & \textbf{76.29\%} & 75.71\% \\
GPT o3 mini (low) & \textbf{78.29\%} & 77.14\% \\
GPT o3 mini (medium) & \textbf{78.57\%} & \textbf{78.57\%} \\
GPT o3 mini (high) & 79.71\% & \textbf{80.00\%} \\ \hline
\end{tabular}
\caption{Accuracy comparison of our framework on the MMLU-PRO dataset with and without Short Phrases. Bold values indicate the better performance for each model. SP denotes Short Phrase.}
  \label{fig:with_short_phrase}
\end{table}

\section{Case Study – BBH-Induct Snarks}

We present a case study on the Snarks (sarcasm detection) task from BBH-Induct. Strategies generated from individual questions are typically much shorter and less detailed compared to those obtained after induction. As a result, applying a single strategy with S-CoT on Gemini 2.0 Flash yields an accuracy of only 0.40. After our induction stage consolidates these strategies, the resulting instruction is more comprehensive and effective (see Figure~\ref{fig:case-study-sarcasm}), leading to a substantial improvement in accuracy to 0.76.

\begin{figure*}[t!]
  \centering
  \scriptsize
  \renewcommand{\arraystretch}{1.1}
  \begin{tabular}{p{15cm}}
    \toprule
    \textbf{\textcolor{blue}{Short Phrase}} \\
    Snarks (Sarcasm Detection)\\
    \midrule
    \textbf{\textcolor{blue}{Question 1}}\\
    (A) Think of the poor man! What will he do without his job? \\
    (B) Think of the poor man! What will he do without his yacht? \\
    \textbf{\textcolor{blue}{Strategy 1}} \\
    Step 1: Analyze option (A) to determine if the statement "Think of the poor man! What will he do without his job?" is likely to be sarcastic. Consider the common understanding of job loss as a serious issue.\\
    
    Step 2: Analyze option (B) to determine if the statement "Think of the poor man! What will he do without his yacht?" is likely to be sarcastic. Consider the common understanding of owning a yacht as a luxury, and the potential for sarcasm when expressing concern about someone losing a luxury item.\\
    
    Step 3: Compare the two options and determine which statement is more likely to be sarcastic based on the context and the implied meaning.\\
    
    Step 4: Select the option (A or B) that is more likely to be sarcastic and output the corresponding letter.\\
    \midrule
    \textbf{\textcolor{blue}{Question 2}}\\
    (A) What a poor guy, years of work and he only got a \$50,000 scholarship \\
    (B) What a poor guy, years of work and he only got a \$50 scholarship \\
    \textbf{\textcolor{blue}{Strategy 2}} \\
    Step 1: Analyze option (A) for potential sarcasm. Consider the implied meaning of "poor guy" in relation to receiving a +\$50,000 scholarship. Evaluate if the statement is intended to convey the opposite of its literal meaning.\\
    
    Step 2: Analyze option (B) for potential sarcasm. Consider the implied meaning of "poor guy" in relation to receiving a \$50 scholarship. Evaluate if the statement is intended to convey the opposite of its literal meaning.\\
    
    Step 3: Compare the two options (A) and (B). Determine which option is more likely to be sarcastic based on the discrepancy between the expected outcome (years of work) and the actual reward (\$50,000 vs. \$50).\\
    
    Step 4: Select the option (A or B) that demonstrates a greater degree of incongruity between the situation and the speaker's expressed sentiment, indicating a higher likelihood of sarcasm.\\
    
    Step 5: Output the corresponding letter (A or B) of the selected option.\\
    \midrule
    \textbf{\textcolor{blue}{Question 3}}\\
    (A) Have you tried not being poor? It is much easier \\
    (B) Have you tried not being rude? It is much easier \\
    \textbf{\textcolor{blue}{Strategy 3}} \\
    Step 1: Analyze option A for sarcastic intent. Consider the inherent difficulty of overcoming poverty and whether the statement implies the opposite.\\
    Step 2: Analyze option B for sarcastic intent. Consider the ease of being polite and whether the statement implies the opposite.\\
    Step 3: Compare the two options and determine which statement is more likely to be interpreted as sarcastic based on common understanding and social context.\\
    Step 4: Select the option (A or B) that exhibits a higher degree of sarcasm.\\
    Step 5: Output the corresponding letter (A or B) as the answer.\\
    \midrule
    \textbf{\textcolor{blue}{Inducted Strategy (STRATEGY-INDUCT)}} \\
    1.  \textbf{Analyze Sentence (A):} Carefully examine sentence (A) for potential indicators of sarcasm. Consider the context and whether the statement could be interpreted as meaning the opposite of what it says, with the intention to criticize or be humorous. Look for elements such as:\\
        \hspace*{1.5em}-- Irony: A statement where the intended meaning is the opposite of the literal meaning.\\
        \hspace*{1.5em}-- Hyperbole: Exaggeration for emphasis or effect.\\
        \hspace*{1.5em}-- Satire: The use of humor, irony, exaggeration, or ridicule to expose and criticize people's stupidity or vices.\\
        \hspace*{1.5em}-- Witty remarks: Cleverly humorous or ironic expressions.\\
        \hspace*{1.5em}-- Ambivalent remarks: Statements that have mixed feelings or contradictory ideas.\\
    
    2.  \textbf{Analyze Sentence (B):} Repeat the analysis process from Step 1 for sentence (B).\\
    
    3.  \textbf{Compare Sentences:} Compare the two sentences based on the likelihood of sarcastic intent. Consider which sentence is more likely to be used to express the opposite of its literal meaning in a critical or humorous way, based on the definition of sarcasm.\\
    
    4.  \textbf{Select the More Sarcastic Sentence:} Determine which sentence (A or B) is more likely to be sarcastic based on your analysis.\\
    
    5.  \textbf{Output the Answer:} Output the letter corresponding to the sentence identified as more likely to be sarcastic.\\
    \bottomrule
  \end{tabular}
  \vspace{-0.7em}
\caption{\small Example strategies and the induced instruction from \textsc{Strategy-Induct} on the snarks (sarcasm detection) task in BBH-Induct (Gemini 2.0 Flash). The induced instruction unifies and formalizes pragmatic cues such as irony and exaggeration from diverse strategies.}
  \label{fig:case-study-sarcasm}
\end{figure*}

\begin{figure*}[ht!]
  \centering
  \footnotesize  
  \begin{tabular}{p{0.9\linewidth}}
    \hline
    You are tasked with designing a strategy to solve a given question. Your goal is to carefully analyze the question and formulate a clear and effective plan outlining the steps needed to solve it. Remember, you should not actually solve the question or provide a final answer.\\[1mm]
    \\
    Additionally, make sure to base your strategy on the details provided in \texttt{<task\_information>}, \texttt{<answer\_format>}, and \texttt{<question>} for the most relevant approach.\\[1mm]
    \\
    Please follow these steps: \\
    1. Carefully read and analyze the question.\\
    2. Identify the key components and challenges within the question.\\
    3. Develop a step-by-step strategy to address the question.\\
    4. Outline your strategy using numbered steps.\\[1mm]
    \\
    Present your strategy in the following format: \\
    \texttt{<strategy>} \\
    Step 1: [Brief description of the first step]\\
    Step 2: [Brief description of the second step]\\
    Step 3: [Brief description of the third step]\\
    \texttt{[Continue with additional steps as needed]}\\
    \texttt{</strategy>}\\[1mm]
    
    **Important:** Focus solely on creating a strategy to solve the question. Do not attempt to solve the question or provide a final answer. Your strategy should outline the approach to solving the question, not the solution itself.\\[2mm]
    \\
    \\
    Here is the task\_information, answer\_format and the question you need to create a strategy for:\\[1mm]
    \texttt{<task\_information>}\\
    \{ \}\\
    \texttt{</task\_information>}\\[1mm]
    
    \texttt{<answer\_format>}\\
    \{ \}\\
    \texttt{</answer\_format>}\\[1mm]
    
    \texttt{<question>}\\
    \{ \}\\
    \texttt{</question>}\\
    \hline
  \end{tabular}
  \caption{Prompt Template for Strategy Generation (Ours).}
  \label{fig:Strategy_Design_Template}
\end{figure*}

\begin{figure*}[ht!]
  \centering
  \footnotesize  
  \begin{tabular}{p{0.9\linewidth}}
    \hline
    You are an expert in the field of NLP (Natural Language Processing), possessing exceptional data observation and analysis skills. Your expertise includes extracting significant rules from complex datasets and formulating precise task instructions based on these rules.\\
    \\
    Currently, you are focusing on analyzing a specific set of examples, deriving insights to formulate a clear and detailed task instruction. This task description will serve as an instruction set, guiding the execution of the related tasks.\\
    \\
    The task instruction should include the following elements: \\
    1. **Task Content**: Clearly define the purpose of the task and the specific activities required to be completed.\\
    2. **Input Format**: Provide detailed descriptions of the types of data accepted, their formats, and how to process these data effectively.\\
    3. **Operational Steps**: Detail the specific step-by-step procedures required to complete the task. Remember to ensure that the final output format adheres to the "answer\_format" specification.\\
    \\
    Ensure that your task instruction is concise, clear, and easily understandable by users. It should provide all necessary information for someone to successfully complete the task.\\
    \\
    Present your task instruction in the following format: \\
    \texttt{<task\_instruction>} \\
    \texttt{\{Your task instruction here, with each element under its own subheading\}} \\
    \texttt{</task\_instruction>} \\
    \\
    Remember to focus solely on creating the task instruction based on the given information and examples. Do not attempt to complete the task itself.\\
    \\
    \\
    Here is the \texttt{task\_information}, \texttt{answer\_format} and the examples you need to create a Task Instruction for: \\
    \texttt{<task\_information>} \\
    \{\} \\
    \texttt{</task\_information>} \\
    \\
    \texttt{<answer\_format>} \\
    \{\} \\
    \texttt{</answer\_format>} \\
    \\
    \texttt{<examples>} \\
    \{\} \\
    \texttt{</examples>} \\
    \hline
  \end{tabular}
  \caption{Prompt Template for INDUCT (Instruction Induction)}
  \label{fig:Task_Instruction_Design_Template}
\end{figure*}

\begin{figure*}[ht!]
  \centering
  \footnotesize  
  \begin{tabular}{p{0.9\linewidth}}
    \hline
    You are an expert in the field of NLP (Natural Language Processing), with outstanding data observation and analysis skills.\\
    \\
    Your expertise includes inductively deriving better solution rules (or methods) from various question-solving strategies and question-pair examples within the same task. Based on these rules, you can formulate precise task instructions.\\
    \\
    Currently, you are focused on analyzing a set of specific question-solving strategies and question-pair examples to gain insights for creating clear and detailed task instruction. This task instruction will serve as instruction sets to guide the execution of related tasks.\\
    \\
    The task instruction should include the following elements: \\
    1. **Task Content**: Clearly define the purpose of the task and the specific activities required to be completed.\\
    2. **Input Format**: Provide detailed descriptions of the types of data accepted, their formats, and how to process these data effectively.\\
    3. **Operational Steps**: Detail the specific step-by-step procedures required to complete the task. Remember to ensure that the final output format adheres to the "answer\_format" specification.\\
    \\
    Ensure your task instruction are concise, clear, and easy to understand for users. They should provide all the necessary information for someone to successfully complete the task.\\
    \\
    Present your task instruction in the following format: \\
    \texttt{<task\_instruction>} \\
    \texttt{\{Your task instruction here, with each element under its own subheading\}} \\
    \texttt{</task\_instruction>} \\
    \\
    Please remember to focus solely on creating the task instruction based on the given information and examples. Do not attempt to complete the task itself.\\
    \\
    \\
    Here is the \texttt{task\_information}, \texttt{answer\_format} and the examples you need to create a Task Instruction for: \\
    \texttt{<task\_information>} \\
    \{\} \\
    \texttt{</task\_information>} \\
    \\
    \texttt{<answer\_format>} \\
    \{\} \\
    \texttt{</answer\_format>} \\
    \\
    \texttt{<examples>} \\
    \{\} \\
    \texttt{</examples>} \\
    \hline
  \end{tabular}
  \caption{Prompt Template for Strategy Induction(Ours)}
  \label{fig:Task_Instruction_Creation_Template}
\end{figure*}

\begin{figure*}[ht!]
  \centering
  \footnotesize  
  \begin{tabular}{p{0.9\linewidth}}
    \hline
    [Task Instruction] \\
    \{\} \\
    \\
    **Output Format**: \\
    \{\} \\
    \\
    Follow these steps carefully: \\
    1. Provide step-by-step deduction that answers the question\\
    \texttt{<deduction>} \\
    \texttt{[Your step-by-step deduction here]} \\
    \texttt{</deduction>} \\\\
    2. Based on your deduction, provide the final answer according to the rules specified in the "Output Format" section.\\
    If unsure and "Output Format" is option, guess the closest option. Present your final answer in the following format:\\
    \texttt{<final\_answer>} \\
    \texttt{[Your final answer here]} \\
    \texttt{</final\_answer>} \\\\
    \textbf{Note} Do not use programming or code to solve this question. \\
    \hline
  \end{tabular}
  \caption{Prompt Template for Zero-shot Chain-of-Thought (ZCoT).}
  \label{fig:BBHI_updated}
\end{figure*}

\begin{figure*}[ht!]
  \centering
  \footnotesize  
  \begin{tabular}{p{0.9\linewidth}}
    \hline
    [Task Instruction] \\
    \{\} \\
    \\
    **Output Format**: \\
    \{\} \\
    \\
    Follow these steps carefully: \\
    1. Carefully consider the problem and generate the strategic knowledge that would best guide the problem-solving process.\\
    \texttt{<strategy>} \\
    \texttt{[Your strategy here]} \\
    \texttt{</strategy>} \\\\
    
    2. Provide step-by-step deduction that answers the question.\\
    \texttt{<deduction>} \\
    \texttt{[Your step-by-step deduction here]} \\
    \texttt{</deduction>} \\\\

    3. Based on your deduction, provide the final answer according to the rules specified in the "Output Format" section.\\
    If unsure and "Output Format" is option, guess the closest option. Present your final answer in the following format:\\
    \texttt{<final\_answer>} \\
    \texttt{[Your final answer here]} \\
    \texttt{</final\_answer>} \\\\
    
    \textbf{Note} Do not use programming or code to solve this question. \\
    \hline
  \end{tabular}
  \caption{Prompt Template for Automatic Strategic Chain-of-Thought (SCoT)}
  \label{fig:ZeroShotCoT_Strategy}
\end{figure*}

\begin{figure*}[ht!]
  \centering
  \footnotesize  
  \begin{tabular}{p{0.9\linewidth}}
    \hline
    [Task Instruction] \\
    \{\} \\
    \\
    **Output Format**: \{\} \\
    \\
    Follow these steps carefully: \\
    1. Provide step-by-step deduction that answers the question\\
    \texttt{<deduction>} \\
    \texttt{[Your step-by-step deduction here]} \\
    \texttt{</deduction>} \\\\
    2. Based on your deduction, provide the final answer according to the rules specified in the "Output Format" section.\\
    If unsure and "Output Format" is option, guess the closest option. Present your final answer in the following format:\\
    \texttt{<final\_answer>} \\
    \texttt{[Your final answer here]} \\
    \texttt{</final\_answer>} \\\\
    \textbf{Note} Do not use programming or code to solve this question. \\
    \hline
  \end{tabular}
  \caption{Prompt Template for INDUCT and \textsc{Strategy-Induct} (Ours)}
  \label{fig:Induct_STRATEGY_INDUCT_Template}
\end{figure*}

\begin{figure*}[ht!]
  \centering
  
  {\scriptsize
  \begin{tabular}{p{0.9\linewidth}}
  \hline
    \textcolor{blue}{Strategy-Induct Prompt} \\
    \texttt{[Task Instruction]} \\[1ex]
    \texttt{\#\#\# Task Content:} \\
    The purpose of this task is to detect sarcasm in a given set of options and select the correct option that exhibits sarcasm. Sarcasm often involves irony, exaggeration, or statements that contradict reality or common expectations. The task requires careful analysis of the tone, context, and implied meaning of the options to identify the sarcastic statement. \\
    \texttt{\#\#\# Input Format:} \\
    - You will be provided with a question containing two or more options (e.g., A, B, etc.). \\
    - Each option is a statement or phrase that needs to be analyzed for sarcasm. \\
    - The input will be presented in a format where each option is labeled with a letter (e.g., A, B). \\
    \texttt{\#\#\# Operational Steps:} \\
    1. \textbf{Understand the Task Requirement}: Recognize that the goal is to detect sarcasm in the given options and select the one that best represents sarcasm. \\[1ex]
    2. \textbf{Analyze the Context of Each Option}: \\
    \quad - Carefully read each option. \\
    \quad - Identify the tone, context, and any potential mismatch between the literal meaning and the implied meaning. \\[1ex]
    3. \textbf{Identify Indicators of Sarcasm}: \\
    \quad - Look for exaggeration, irony, or statements that seem to contradict reality or common expectations. \\
    \quad - Pay attention to phrases that imply the opposite of what is being said or highlight an absurdity. \\[1ex]
    4. \textbf{Compare the Options}: \\
    \quad - Evaluate each option to determine which one contains a sarcastic tone. \\
    \quad - Consider the likelihood of the statement being ironic, exaggerated, or mocking. \\[1ex]
    5. \textbf{Select the Correct Option}: \\
    \quad - Choose the single letter (e.g., A, B, etc.) corresponding to the option that best represents sarcasm. \\[1ex]
    6. \textbf{Verify the Choice}: \\
    \quad - Double-check your reasoning to ensure the selected option aligns with the task of sarcasm detection. \\[1ex]
    7. \textbf{Output the Answer}: \\
    \quad - Provide the output as a single letter (e.g., A, B, etc.) corresponding to the chosen option, adhering to the specified answer format. \\
    \\
    \textcolor{blue}{Prompt Template}\\
    \textbf{Output Format}: \\
    The output is a single letter corresponding to the correct option from the list (e.g., A, B, C, etc.). \\
    Follow these steps carefully: \\
    1. Provide step-by-step deduction that answers the question \\
    \quad <deduction> \\
    \quad [Your step-by-step deduction here] \\
    \quad </deduction> \\
    2. Based on your deduction, provide the final answer according to the rules specified in the "Output Format" section. \\
    If unsure and "Output Format" is option, guess the closest option. Present your final answer in the following format: \\
    \quad <final\_answer> \\
    \quad [Your final answer here] \\
    \quad </final\_answer> \\
    \\ 
    \textcolor{blue}{New Question}\\ 
    \texttt{[Question]} \\ 
    (A) The best way to defeat a stronger/larger and better regimented force is by fighting on their terms \\
    (B) The best way to defeat a stronger/larger and better regimented force is by fighting on your terms \\
    \texttt{[Answer]} \\ 
    Let's refer to the "Operational Steps" and think step by step. \\ 
    \\
    \\ 
    \textcolor{blue}{Model Response}\\ 
    \texttt{<deduction>} \\ 
    1. \textbf{Understand the Task Requirement}: The task is to detect sarcasm in the given options and select the one that best represents sarcasm. Sarcasm often involves irony, exaggeration, or a contradiction to reality or common expectations. \\ 
    2. \textbf{Analyze the Context of Each Option}: \\ 
    \quad - \textbf{Option A}: "The best way to defeat a stronger/larger and better regimented force is by fighting on their terms." This statement suggests that one should fight a stronger and better-organized force by adhering to their strengths and strategies. This seems counterintuitive and ironic because it contradicts common sense; fighting on the enemy's terms would likely lead to defeat. \\ 
    \quad - \textbf{Option B}: "The best way to defeat a stronger/larger and better regimented force is by fighting on your terms." This statement aligns with common sense and strategy. It suggests that one should fight in a way that plays to their own strengths, which is logical and not sarcastic. \\ 
    3. \textbf{Identify Indicators of Sarcasm}: \\ 
    \quad - \textbf{Option A} contains irony because it suggests doing something that is clearly counterproductive (fighting on the enemy's terms). This is a sarcastic way of pointing out the absurdity of such an approach. \\ 
    \quad - \textbf{Option B} is straightforward and lacks any ironic or sarcastic tone. \\ 
    4. \textbf{Compare the Options}: \\ 
    \quad - Option A is sarcastic because it uses irony to highlight the absurdity of fighting on the enemy's terms. \\ 
    \quad - Option B is not sarcastic; it is a logical and straightforward statement. \\ 
    5. \textbf{Select the Correct Option}: \\ 
    \quad - Based on the analysis, Option A is the sarcastic statement. \\ 
    6. \textbf{Verify the Choice}: \\ 
    \quad - The reasoning aligns with the task of sarcasm detection. Option A uses irony to make its point, making it the correct choice. \\ 
    </deduction> \\ 
    <final\_answer> \\ 
    \textcolor{green!30!black}{A}\\
    </final\_answer> \\
    \hline
  \end{tabular}
  }
\caption{Example of Strategy-Induct Prompting on the \textit{Snarks} task in the BBH-Induct dataset, using GPT-4o, where the correct answer is A.}
  \label{tab:StrategyInductPrompting}
\end{figure*}

\begin{figure*}[ht!]
  \centering
  {\scriptsize  
  \begin{tabular}{p{0.9\linewidth}}
  \hline
    \textcolor{blue}{Zero Shot CoT Prompt} \\
    \texttt{[Task Instruction]} \\
    Snarks (Sarcasm Detection) \\[1ex]
    \\
    \textcolor{blue}{Prompt Template} \\
    \textbf{Output Format}: The output is a single letter corresponding to the correct option from the list (e.g., A, B, C, etc.). \\

    Follow these steps carefully: \\
    1. Provide step-by-step deduction that answers the question \\
    \quad \texttt{<deduction>} \\
    \quad [Your step-by-step deduction here] \\
    \quad \texttt{</deduction>} \\

    2. Based on your deduction, provide the final answer according to the rules specified in the "Output Format" section. \\
    If unsure and "Output Format" is option, guess the closest option. Present your final answer in the following format: \\
    \quad \texttt{<final\_answer>} \\
    \quad [Your final answer here] \\
    \quad \texttt{</final\_answer>} \\

    \textbf{Note}: Do not use programming or code to solve this question. \\
    \\
    \textcolor{blue}{New Question}\\ 
    \texttt{[Question]} \\
    (A) The best way to defeat a stronger/larger and better regimented force is by fighting on their terms \\
    (B) The best way to defeat a stronger/larger and better regimented force is by fighting on your terms \\

    [Answer] \\
    Let's think step by step. \\

    \\
    \textcolor{blue}{Model Response}\\  
    \texttt{<deduction>} \\
    1. The question revolves around the strategy to defeat a stronger, larger, and better-regimented force. This implies that the opponent has an advantage in terms of size, strength, and organization. \\
    2. Option (A) suggests fighting on the opponent's terms. This would mean engaging in a way that plays to the opponent's strengths, which is counterintuitive. If the opponent is stronger and better organized, fighting on their terms would likely lead to defeat. \\
    3. Option (B) suggests fighting on your own terms. This implies leveraging your own strengths and avoiding the opponent's advantages. Historically and strategically, this approach is more effective when facing a superior force. For example, guerrilla warfare or asymmetric tactics are often used to counteract a stronger opponent. \\
    4. Based on logic and historical precedent, fighting on your own terms (Option B) is the more effective strategy. \\
    \texttt{</deduction>} \\

    \texttt{<final\_answer>} \\
    \textcolor{red}{B} \\
    \texttt{</final\_answer>} \\
    \hline
  \end{tabular}
  }
\caption{Example of Zero-shot CoT Prompting on the \textit{Snarks} task in the BBH-Induct dataset, using GPT-4o, where the correct answer is A.}

  \label{tab:SnarksSarcasmDetection}
\end{figure*}

\clearpage

\begin{table*}[h]
\centering
\small
\begin{tabular}{c|c|c}
\hline
\textbf{Task} & \textbf{Sub Task} & \textbf{Short Phrase} \\ \hline
Boolean Expressions & boolean expressions & Boolean Expressions \\ \hline
Causal Judgment & causal judgement & Causal Judgment \\ \hline
Date Understanding & date understanding & Date Understanding \\ \hline
Disambiguation QA & disambiguation qa & Disambiguation QA \\ \hline
Dyck Languages & dyck languages & Dyck Languages \\ \hline
Formal Fallacies Syllogisms Negation & formal fallacies & Formal Fallacies Syllogisms Negation \\ \hline
Geometric Shapes & geometric shapes & Geometric Shapes \\ \hline
Hyperbaton (Adjective Ordering) & hyperbaton & Hyperbaton (Correct Adjective Order) \\ \hline
\multirow{3}{*}{Logical Deduction} & logical deduction five objects & Logical Deduction \\ \cline{2-3} 
 & logical deduction seven objects & Logical Deduction \\ \cline{2-3} 
 & logical deduction three objects & Logical Deduction \\ \hline
Movie Recommendation & movie recommendation & Similar Movie Recommendation \\ \hline
Multi-Step Arithmetic & multistep arithmetic two & Multi-Step Arithmetic \\ \hline
Navigate & navigate & Navigate (Round-Trip Check) \\ \hline
Object Counting & object counting & Total Object Count \\ \hline
Penguins in a Table & penguins in a table & Penguins in a Table \\ \hline
Reasoning about Colored Objects & reasoning about colored objects & Reasoning about Colored Objects \\ \hline
Ruin Names & ruin names & Ruin Names (Humorous Edit) \\ \hline
Salient Translation Error Detection & salient translation error detection & Salient Translation Error Detection \\ \hline
Snarks & snarks & Snarks (Sarcasm Detection) \\ \hline
Sports Understanding & sports understanding & Sports Plausibility Check \\ \hline
Temporal Sequences & temporal sequences & Temporal Sequences \\ \hline
\multirow{3}{*}{Tracking Shuffled Objects} & tracking shuffled objects five objects & Tracking Shuffled Objects \\ \cline{2-3} 
 & tracking shuffled objects seven objects & Tracking Shuffled Objects \\ \cline{2-3} 
 & tracking shuffled objects three objects & Tracking Shuffled Objects \\ \hline
Web of Lies & web of lies & Web of Lies \\ \hline
Word Sorting & word sorting & Word Sorting \\ \hline
\end{tabular}
\caption{Mapping between tasks, subtasks, and their corresponding short phrases in BBH-Induct.}
\label{tab:task_subtask_shortphrase}
\end{table*}

\begin{table*}[h!]
\centering
\small
\begin{tabular}{c|c|c}
\hline
\textbf{Task} & \textbf{Sub Task} & \textbf{Short Phrase} \\ \hline
2d\_movement & - & 2D Movement (Facing North) \\ \hline
anagrams & - & Anagram Solver \\ \hline
bitwise & - & Bitwise Arithmetic \\ \hline
count\_token\_freq\_dna & - & DNA Token Frequency Counter \\ \hline
\multirow{2}{*}{css-selectors} & css-selectors\_explain & CSS Selectors Verifier \\ \cline{2-3}
 & css-selectors\_verbal & CSS Selector Generator \\ \hline
determinant & - & Matrix Determinant Calculator \\ \hline
forth\_stack\_sim & - & Forth Stack Simulator \\ \hline
guess\_the\_singer & - & Guess the Singer \\ \hline
largest\_country & - & Largest Country \\ \hline
lat\_long\_identify & - & Lat-Long Country Finder \\ \hline
mate-in-one & - & Mate in One \\ \hline
math\_equations & - & Math Equations \\ \hline
missing\_operators & - & Missing Operators \\ \hline
ner\_finance & - & NER Finance \\ \hline
next-val-series & - & Next-Val Series \\ \hline
partially\_solved\_crossword\_clues & - & Partially Solved Crossword Clues \\ \hline
points\_on\_line & - & Midpoint on Line \\ \hline
poker\_analysis & - & Poker Winner \\ \hline
recurrence-relation & - & Recurrence Relation \\ \hline
resistor\_ohm\_calculator & - & Resistor Ohm Calculator \\ \hline
smiles\_to\_formula & - & SMILES to Formula \\ \hline
sort\_numeric & - & Sort Numeric Descending \\ \hline
\multirow{4}{*}{word\_association} & word\_association\_related\_words\_2 & Word Association \\ \cline{2-3}
 & word\_association\_related\_words\_3 & Word Association \\ \cline{2-3}
 & word\_association\_related\_words\_4 & Word Association \\ \cline{2-3}
 & word\_association\_related\_words\_5 & Word Association \\ \hline
\end{tabular}
\caption{Mapping between tasks, subtasks, and their corresponding short phrases in Evals-Induct.}
\label{tab:EI_task_mapping}
\end{table*}

\begin{table*}[h!]
\centering
\small
\begin{tabular}{>{\centering\arraybackslash}p{3cm} | >{\centering\arraybackslash}p{4cm} | >{\centering\arraybackslash}p{3cm}}

\hline
\textbf{Task} & \textbf{Sub Task} & \textbf{Short Phrase} \\ \hline
\multirow{26}{*}{Shift Cipher} 
& Shift Cipher – ROT-1  & \multirow{26}{*}{Shift Cipher} \\ \cline{2-2}
& Shift Cipher – ROT-2  &  \\ \cline{2-2}
& Shift Cipher – ROT-3  &  \\ \cline{2-2}
& Shift Cipher – ROT-4  &  \\ \cline{2-2}
& Shift Cipher – ROT-5  &  \\ \cline{2-2}
& Shift Cipher – ROT-6  &  \\ \cline{2-2}
& Shift Cipher – ROT-7  &  \\ \cline{2-2}
& Shift Cipher – ROT-8  &  \\ \cline{2-2}
& Shift Cipher – ROT-9  &  \\ \cline{2-2}
& Shift Cipher – ROT-10 &  \\ \cline{2-2}
& Shift Cipher – ROT-11 &  \\ \cline{2-2}
& Shift Cipher – ROT-12 &  \\ \cline{2-2}
& Shift Cipher – ROT-13 &  \\ \cline{2-2}
& Shift Cipher – ROT-14 &  \\ \cline{2-2}
& Shift Cipher – ROT-15 &  \\ \cline{2-2}
& Shift Cipher – ROT-16 &  \\ \cline{2-2}
& Shift Cipher – ROT-17 &  \\ \cline{2-2}
& Shift Cipher – ROT-18 &  \\ \cline{2-2}
& Shift Cipher – ROT-19 &  \\ \cline{2-2}
& Shift Cipher – ROT-20 &  \\ \cline{2-2}
& Shift Cipher – ROT-21 &  \\ \cline{2-2}
& Shift Cipher – ROT-22 &  \\ \cline{2-2}
& Shift Cipher – ROT-23 &  \\ \cline{2-2}
& Shift Cipher – ROT-24 &  \\ \cline{2-2}
& Shift Cipher – ROT-25 &  \\ \hline
\end{tabular}
\caption{Mapping between tasks, subtasks, and their corresponding short phrases in Shift Cipher.}
\label{tab:SC_task_mapping}
\end{table*}

\end{document}